\documentclass{article}

\usepackage{arxiv}

\usepackage[utf8]{inputenc} 
\usepackage[T1]{fontenc}    
\usepackage[backref=page]{hyperref}       
\usepackage{url}            
\usepackage{booktabs}       
\usepackage{amsfonts}       
\usepackage{nicefrac}       
\usepackage{microtype}      
\usepackage[bottom]{footmisc}
\usepackage{graphicx}
\usepackage{subfigure}
\usepackage{booktabs} 
\usepackage{amsmath,amsfonts}
\usepackage{xcolor}
\usepackage{amsthm}
\usepackage{verbatim}
\usepackage{algorithm}
\usepackage{algorithmic}
\usepackage{comment}

\def\~#1{\mathbb{#1}}
\def\*#1{\mathbf{#1}}
\newcommand{\mcal}[1]{\mathbf{\mathcal{#1}}}
\newcommand\norm[1]{\left\lVert#1\right\rVert}

\def\oscore{{\texttt{Score}}}
\def\online{{\texttt{LineFilter}}}

\def\kernelfilter{{\texttt{KernelFilter}}}
\def\mrlf{{\texttt{StreamingLF}}}
\def\mrlw{{\texttt{StreamingLW}}}
\def\mrwcb{{\texttt{StreamingWCB}}}
\def\mrfc{{\texttt{StreamingFC}}}
\def\uni{{\texttt{Uniform}}}
\def\stream{{\texttt{Streaming}}}
\def\streamXX{{\texttt{StreamingXX}}}

\newtheorem{definition}{Definition}[section]
\newtheorem{theorem}{Theorem}[section]
\newtheorem{corollary}{Corollary}[section]
\newtheorem{lemma}{Lemma}[section]

\title{Streaming Coresets for Symmetric Tensor Factorization}

\author{
  Rachit Chhaya\\
  IIT Gandhinagar\\
  \texttt{rachit.chhaya@iitgn.ac.in} \\
   \And
  Jayesh Choudhari\\
  IIT Gandhinagar\\
  \texttt{choudhari.jayesh@iitgn.ac.in} \\
   \And
 Anirban Dasgupta \\
  IIT Gandhinagar\\
  \texttt{anirbandg@iitgn.ac.in} \\
  \AND
   Supratim Shit \thanks{Corresponding author} \\
  IIT Gandhinagar \\
  \texttt{supratim.shit@iitgn.ac.in} \\
}

\begin{document}
\maketitle

\begin{abstract}
Factorizing tensors has recently become an important optimization module in a number of machine learning pipelines, especially in latent variable models. We show how to do this efficiently in the streaming setting. Given a set of $n$ vectors, each in $\~R^d$, we present algorithms to select a sublinear number of these vectors as coreset, while guaranteeing that the CP decomposition of the $p$-moment tensor of the coreset approximates the corresponding decomposition of the $p$-moment tensor computed from the full data. We introduce two novel algorithmic techniques: online filtering and kernelization. Using these two, we present six algorithms that achieve different tradeoffs of coreset size, update time and working space, beating or matching various state of the art algorithms. In the case of matrices ($2$-ordered tensor), our online row sampling algorithm guarantees $(1 \pm \epsilon)$ relative error spectral approximation. We show applications of our algorithms in learning single topic modeling. 
\end{abstract}


\backrefsetup{enable}
\allowdisplaybreaks

\section{Introduction}
Much of the data that is consumed in data mining and machine learning applications arrives in a streaming manner. The data is conventionally treated as a matrix, with a row representing a single data point and the columns its corresponding features. Since the matrix is typically large, it is advantageous to be able to store only a small number of rows and still preserve some of its ``useful" properties. One such abstract property that has proven useful in a number of different settings, such as solving regression, finding various factorizations, is {\em subspace preservation}. Given a matrix $\*A \in \~R^{n \times d}$, an $m \times d$ matrix $\*C$ is its subspace preserving matrix for the $\ell_2$ norm if, $\forall \*x \in \~R^{d}$,
\begin{align*}
    \big| \sum_{\tilde{\*a}_{j} \in \*C}(\tilde{\*a}_{j}^T \*x)^2 - \sum_{i \in [n]}(\*a_{i}^T \*x)^2\big|\leq\epsilon \cdot \sum_{i \in [n]}(\*a_{i}^T \*x)^2
\end{align*}
We typically desire $m \ll n$ and $\tilde{\*a}_{j}$'s represent the subsampled and rescaled rows from $\*A$. Such a sample $\*C$ is often referred to as a coreset.
This property has been used to obtain approximate solutions to many problems such as regression, low-rank approximation, etc~\cite{woodruff2014sketching} while having $m$ to be at most $O(d^2)$. Such property has been defined for other $\ell_p$ norms too \cite{dasgupta2009sampling,cohen2015p,clarkson2016fast}. 

Matrices are ubiquitous, and depending on the application, one can assume that the data is coming from a generative model, i.e., there is some distribution from which every incoming data point (or row) is sampled and given to user. Many a time, the goal is to know the hidden variables of this generative model. An obvious way to learn these variables is by representing data (matrix) by its low-rank representation. However, we know that a low-rank representation of a matrix is not unique as there are various ways (such as SVD, QR, LU)  to decompose a matrix. So it difficult to realize the hidden variables just by the low-rank decomposition of the data matrix.
This is one of the reasons to look at higher order moments of the data i.e. tensors. Tensors are formed by outer product of data vectors, i.e. for a dataset $\*A \in \~R^{n \times d}$ one can use a $p$ order tensor $\mcal T \in \~R^{d \times \ldots \times d}$ as $\mcal T = \sum_{i=1}^{n} \*a_{i} \otimes^{p}$, where $p$ is set by user depending on the number of latent variables one is expecting in the generative model~\cite{ma2016polynomial}. The decomposition of such a tensor is unique under a mild assumption~\cite{kruskal1977three}. 
Factorization of tensors into its constituent elements has found uses in many machine learning applications such as topic modeling~\cite{anandkumar2014tensor}, various latent variable models~\cite{anandkumar2012method,hsu2012spectral,jenatton2012latent}, training neural networks~\cite{janzamin2015beating} etc. 

For a $p$-order moment tensor $\mcal T = \sum_i \*a_{i}\otimes^p$ created using the set of vectors $\{\*a_{i}\}$ and for $\*x \in \~R^{d}$ one of the important property one needs to preserve is $\mcal T(\*x, _{\cdots}, \*x) = \sum_{i} (\*a_{i}^{T}\*x)^{p}$. This operation is also called {\em tensor contraction} \cite{song2016sublinear}. Now if we wish to ``approximate" it using only a subset of the rows in $\*A$, the above property for $\ell_2$ norm subspace preservation does not suffice. What suffices, however, is a guarantee that is similar (not same) to that needed for the $\ell_p$ subspace preservation.  

For tensor factorization, which is performed using power iteration, a coreset $\*C \subseteq \{\*a_{i}\}$, in order to give a guaranteed approximation to the tensor factorization, needs to satisfy the following natural extension of the $\ell_2$ subspace preservation condition:
\begin{align*}
\sum_{\*a_{i} \in \*A}({\*a_{i}}^T \*x)^p \approx \sum_{\tilde{\*a}_{j} \in \*C}({\tilde{\*a}_{j}}^T \*x)^p
\end{align*}
 Ensuring this tensor contraction property enables one to approximate the CP decomposition of $\mcal T$ using only the vectors $\tilde{\*a}_{i}$'s via power iteration method~\cite{anandkumar2014tensor}. A related notion is that of $\ell_p$ subspace embedding where we need that $\*C$ satisfies the following, $\forall \*x\in \~R^{d}$
\begin{align*}
\sum_{\*a_{i} \in \*A}|{\*a_{i}}^T \*x|^p \approx \sum_{\tilde{\*a}_{j} \in \*C}|{\tilde{\*a}_{j}}^T \*x|^p
\end{align*}
This property ensures that we can approximate the $\ell_p$ regression problem by using only the rows in $\*C$. 
The two properties are the same for even $p$, as both LHS and RHS is just the sum of non-negative terms. But they slightly differ for odd values of $p$.

In this work, we show that it is possible to create coresets for the above property in streaming and restricted streaming ({\em online}) setting. In restricted streaming setting an incoming point, when it arrives, is either chosen in the set or discarded forever. We consider the following formalization of the above two properties. Given a query space of vectors $\*Q\subseteq \~R^{d}$ and $\epsilon > 0$, we aim to choose a set $\*C$ which contains sampled and rescaled rows from $\*A$ to ensure that $\forall \*x \in \*Q$ with probability at least $0.99$, the following properties hold,
\begin{align}
    \big|\sum_{\tilde{\*a}_{j} \in \*C}(\tilde{\*a}_{j}^T\*x)^p - \sum_{i \in [n]}(\*a_{i}^T \*x)^p\big| \leq \epsilon \cdot \sum_{i \in [n]}|\*a_{i}^T \*x|^p  \label{eq:contract} \\
    \big|\sum_{\tilde{\*a}_{j} \in \*C}|\tilde{\*a}_{j}^T\*x|^p - \sum_{i \in [n]}|\*a_{i}^T \*x|^p\big| \leq \epsilon \cdot \sum_{i \in [n]}|\*a_{i}^T \*x|^p  \label{eq:lp}
\end{align}
Note that neither property follows from the other. For even values of $p$, the above properties are identical and imply a relative error approximation as well. For odd values of $p$, the $\ell_{p}$ subspace embedding as equation \eqref{eq:lp} gives a relative error approximation but the tensor contraction as equation \eqref{eq:contract} implies an additive error approximation, as LHS terms are not sum of absolute terms. It can become relative error under non-negativity constraints on $\*a_{i}$ and $\*x$. This happens, for instance, for the important use case of topic modeling, where $p=3$ typically.
%
\begin{table*}[t]
\caption{\small{Table comparing existing work (first four rows) and current contributions. \streamXX~refers to the obvious extension of the \texttt{XX} algorithm to the streaming model using merge-reduce.}}
\label{tab:compare}
\vskip 0.1in
\begin{center}
\begin{scriptsize}
\begin{sc}
\begin{tabular}{|l|l|l|l|}
\hline
Algorithm & Sample Size $\tilde{O}(\cdot)$  & Update time   & Working space $\tilde{O}(\cdot)$\\
\hline\hline
\mrwcb~\cite{dasgupta2009sampling} & $d^{p}k\epsilon^{-2}$  & $d^{5}p\log d$ amortized & $d^{p}k\epsilon^{-2}$ \\
\hline
\mrlw~\cite{cohen2015p} & $d^{p/2}k\epsilon^{-5}$  & $d^{p/2}$ amortized & $d^{p/2}k\epsilon^{-5}$ \\
\hline
\mrfc~\cite{clarkson2016fast} & $d^{7p/2}\epsilon^{-2}$  & $d$ amortized & $d^{7p/2}\epsilon^{-2}$ \\
\hline
\stream~\cite{dickens2018leveraging} & $n^{\gamma}d\epsilon^{-2}$  & $n^{\gamma}d^{5}$ & $n^{\gamma}d$ \\
\hline
\hline
\online~(Theorem~\ref{thm:Online}) & $n^{1-2/p}dk\epsilon^{-2}$ & $d^{2}$ & $d^2$ \\
\hline
\online+\mrlw~(Theorem~\ref{thm:improvedStream-MR}) & $d^{p/2}k\epsilon^{-5}$ & $d^{2}$ amortized & $d^{p/2}k\epsilon^{-5}$ \\
\hline
\kernelfilter~(Theorem~\ref{thm:slowOnline})(even $p$) & $d^{p/2}k\epsilon^{-2}$ & $d^{p}$ & $d^{p}$ \\
\hline
\kernelfilter~(Theorem~\ref{thm:slowOnlineOdd})(odd $p$) & $n^{1/(p+1)}d^{p/2}k\epsilon^{-2}$ & $d^{p+1}$ & $d^{p+1}$ \\
\hline
\online+\kernelfilter~(Theorem~\ref{thm:improvedOnlineCoreset})(even $p$) & $d^{p/2}k\epsilon^{-2}$ & $d^{2}$ amortized & $d^{p}$ \\
\hline
\online+\kernelfilter~(Theorem~\ref{thm:improvedOnlineCoreset})(odd $p$) & $n^{{(p - 2)}/{(p^2+p)}}d^{p/2+1/4}k^{5/4}\epsilon^{-2}$ & $d^{2}$ amortized & $d^{p+1}$ \\
\hline  
\online+\mrlw+\kernelfilter~(Theorem~\ref{thm:lflwkf}) & $d^{p/2+1/2}k^{5/4}\epsilon^{-2}$ & $d^{2}$ amortized & $d^{p+1}$ \\
\hline  
\end{tabular}
\end{sc}
\end{scriptsize}
\end{center}
\vskip -0.1in
\end{table*}
\paragraph{Our Contributions:}
We propose various methods to sample rows in streaming manner for a $p$ order tensor, which is further decomposed to know the latent factors. For a given matrix $\*A \in \~R^{n\times d}$, a $k$-dimensional query space $\*Q \in \~R^{k \times d},\epsilon>0$ and $p\geq 2$,
\begin{itemize}
    \item We give an algorithm (\online) that is able to select rows, it takes $O(d^2)$ update time and working space to return a sample of size $O\Big(\frac{n^{1-2/p} dk}{\epsilon^{2}} \big(1+\log\|\*A\| - d^{-1}\min_{i}\log\|\*a_{i}\|\big)\Big)$ such that the set of selected rows forms a coreset having the guarantees stated in equations \eqref{eq:contract} and \eqref{eq:lp} (Theorem~\ref{thm:Online}). It is a streaming algorithm but also works well in the restricted streaming (online) setting.
    \item We improve the sampling complexity of our coreset to $O(d^{p/2}k(\log n)^{10}\epsilon^{-5})$ by a streaming algorithm (\online+\mrlw) with amortized update time $O(d^2)$ (Theorem~\ref{thm:improvedStream-MR}). It requires slightly higher working space $O(d^{p/2}k(\log n)^{11}\epsilon^{-5})$.
    \item For integer value $p \geq 2$ we present a kernelization technique which, for any vector $\*v$, uses two vectors $\grave{\*v}$ and $\acute{\*v}$ such that for any $\*x, \*y \in \~R^{d}$,
    \begin{equation*}
        |\*x^T \*y|^p = |\grave{\*x}^T \grave{\*y}|\cdot|\acute{\*x}^T \acute{\*y}|
    \end{equation*}
    Using this technique, we give an algorithm (\kernelfilter) which takes $O(nd^{p})$ time and samples $O\Big(\frac{d^{p/2}k}{\epsilon^{2}} \big(1+p(\log\|\*A\| - d^{-p/2}\min_{i}\log\|\*a_{i}\|\big)\Big)$ vectors to create a coreset having the same guarantee as~\eqref{eq:contract} and \eqref{eq:lp} (Theorem~\ref{thm:slowOnline}) for even value $p$. For odd value $p$ it takes $O(nd^{p+1})$ time and samples $O\Big(\frac{n^{1/(p+1)}d^{p/2}k}{\epsilon^{2}}\big(1+(p+1)(\log\|\*A\| - d^{-\lceil p/2 \rceil}\min_{i}\log\|\*a_{i}\|)\big)^{p/(p+1)}\Big)$ vectors to create a coreset having the same guarantee as~\eqref{eq:contract} and \eqref{eq:lp} (Theorem~\ref{thm:slowOnlineOdd}).
    Both update time and working space of the algorithm for even $p$ is $O(d^{p})$ and for odd $p$ it is $O(d^{p+1})$. It is a streaming algorithm but also works well in the restricted streaming (online) setting.
    \item For integer value $p \geq 2$ we combine both the online algorithms and propose another online algorithm (\online+\kernelfilter) which has $O(d^2)$ amortized update time and returns $O\Big(\frac{d^{p/2}k}{\epsilon^{2}} \big(1+p(\log\|\*A\| - d^{-p/2}\min_{i}\log\|\*a_{i}\|\big)\Big)$ for even $p$ and $O\Big(\frac{n^{(p-2)/(p^{2}+p)}d^{p/2+1/4}k^{5/4}}{\epsilon^{2}}\big(1+(p+1)(\log\|\*A\| - d^{-\lceil p/2 \rceil}\min_{i}\log\|\*a_{i}\|)\big)^{p/(p+1)}\Big)$ vectors for odd $p$ as coreset with same guarantees as equation~\eqref{eq:contract} and ~\eqref{eq:lp} (Theorem~\ref{thm:improvedOnlineCoreset}). Here the working space is same as \kernelfilter. The factor $n^{(p-2)/(p^{2}+p)} \leq n^{1/10}$. 
    \item We also propose a streaming algorithm (\online+\mrlw+\kernelfilter) specially for odd value $p$. The algorithm takes $O(d^2)$ amortized update time and returns $O\Big(\frac{d^{p/2+1/2}k^{5/4}}{\epsilon^{2}}\big(1+(p+1)(\log\|\*A\| - d^{-\lceil p/2 \rceil}\min_{i}\log\|\*a_{i}\|)\big)^{p/(p+1)}\Big)$ vectors as coreset with same guarantees as equation~\eqref{eq:contract} and ~\eqref{eq:lp} (Theorem~\ref{thm:lflwkf}). The working space is same as \kernelfilter, i.e., $O(d^{p+1})$. 
    \item We give a streaming algorithm (\mrlf), which is a streaming version of \online. It takes $\tilde{O}(d_{\zeta})$ update time, which is linear in the dimension of input rows and $O(n^{1-2/p}dk(\log n)^{5}\epsilon^{-2})$ working space to return a coreset of size $O\big(\frac{n^{1-2/p} dk(\log n)^{4}}{\epsilon^{2}}\big)$ such that the set of selected rows forms a coreset having the guarantees stated in equations \eqref{eq:contract} and \eqref{eq:lp} (Theorem~\ref{thm:stream-LF}). Unlike \online which can be used in both steaming and online setting, \mrlf~is a streaming algorithm.
    \item For the $p=2$ case, both \online~and \kernelfilter~translate to an online algorithm for sampling rows of the matrix $\*A$, while guaranteeing a {\em relative error} spectral approximation (Theorem~\ref{thm:improvedMatrixCoreset}). 
    This is an improvement (albeit marginal) over the online row sampling result by~\cite{cohen2016online}. The additional benefit of this new online algorithm over~\cite{cohen2016online} is that it does not need knowledge of $\sigma_{\min}(\*A)$ to give a relative error approximation. 
\end{itemize}
The rest of this paper is organized as follows: In section \ref{sec:prelimnary}, we look at some preliminaries for tensors and coresets. We also describe the notation used throughout the paper. Section \ref{sec:related} discusses related work. In section \ref{sec:algorithms}, we state all the six streaming algorithms along with their guarantees. We also show how our problem of preserving tensor contraction relates to preserving $\ell_{p}$ subspace embedding. In section \ref{sec:proofs}, we describe the guarantees given by our algorithm and their proofs. In section \ref{sec:topic}, we describe how our algorithm can be used in case of streaming single topic modeling. We give empirical results that compare our sampling scheme with other schemes.
\section{Preliminaries}{\label{sec:prelimnary}}
We use the following notation throughout the paper. A scalar is denoted by a lower case letter, e.g., $p$ while a vector is denoted by a boldface lower case letter, e.g., $\*a$. 
By default, all vectors are considered as column vectors unless specified otherwise. Matrices and sets are denoted by boldface upper case letters, e.g., $\*A$. 
Specifically, $\*A$ denotes an $n\times d$ matrix
with set of rows $\{\*a_{i}\}$ and, in the streaming setting, $\*A_{i}$ represents the matrix formed by the first $i$ rows of $\*A$ that have arrived. We will interchangeably refer to the set $\{\*a_i\}$ as the input set of vectors as well as the rows of the matrix $\*A$. A tensor is denoted by a bold calligraphy letter e.g. $\mcal T$. Given a set of $d-$dimensional vectors $\*a_1, \ldots, \*a_n$, from which a $p$-order symmetric tensor $\mcal T$ is obtained as
$ \mcal T = \sum_{i=1}^n \*a_i \otimes^{p}$
i.e., the sum of the $p$-order outer product of each of the vectors. It is easy to see that $\mcal T$ is a symmetric tensor as it satisfies
the following: $\forall i_{1},i_{2},_{\cdots},i_{p}; \mcal T_{i_{1},i_{2},_{\cdots},i_{p}} = \mcal T_{i_{2},i_{1},_{\cdots},i_{p}} = _{\cdots} = \mcal T_{i_{p},i_{p-1},_{\cdots},i_{1}}$, i.e. all the tensor entries with indices as some permutations of $(i_1, i_2, _{\cdots}, i_p)$ have the same value. We define the scalar quantity, also known as tensor contraction, as $\mcal T(\*x,\ldots,\*x) = \sum_{i=1}^{n}(\*a_i^T\*x)^p$, where $\*x \in \~R^{d}$. There are three widely used tensor decomposition techniques known as CANDECOMP/PARAFAC(CP), Tucker and Tensor Train decomposition \cite{kolda2009tensor, oseledets2011tensor}. Our work focuses on CP decomposition. In rest of the paper tensor decomposition is referred as CP decomposition.

We denote 2-norm for a vector $\*x$ as $\|\*x\|$, and any $p$-norm, for $p\neq 2$ as $\|\*x\|_p$. We denote the 2-norm or spectral norm of a matrix $\*A$ by $\|\*A\|$. For a $p$-order tensor $\mcal T$ we denote the spectral norm as $\|\mcal T\|$ which is defined as $\|\mcal T\| = \sup_{\*x}\frac{|\mcal T(\*x,\ldots,\*x)|}{\|\*x\|^{p}}$.

\textbf{Coreset:}
It is a small summary of data which can give provable guarantees for a particular optimization problem. Formally, given a set
$\*X \subseteq \~R^d$, query set $\*Q$, a non-negative cost function $\mathnormal{f}_{\*q}(\*x)$ with parameter $\*q \in \*Q$ and data point $\*x \in \*X$, a set of subsampled and appropriately reweighed points $\*C$ is called a {\em coreset} if $\forall \*q \in \*Q$,  $|\sum_{\*x \in \*X}\mathnormal{f}_{\*q}(\*x) - \sum_{\tilde{\*x} \in \*C}\mathnormal{f}_{\*q}(\mathbf{\tilde{x}})| \leq \epsilon\sum_{\*x \in \*X}\mathnormal{f}_{\*q}(\*x)$ for some $\epsilon > 0$.
We can relax the definition of coreset to allow a small additive error. For $\epsilon, \gamma > 0$, we can have a subset $\*C\subseteq \*X$ such that $\forall \*q \in \*Q$, $|\sum_{\*x \in \*X}\mathnormal{f}_{\*q}(\*x) - \sum_{\mathbf{\tilde{x}} \in \*C}\mathnormal{f}_{\*q}(\mathbf{\tilde{x}})| \leq \epsilon\sum_{\*x \in \*X}\mathnormal{f}_{\*q}(\*x) + \gamma$. 

To guarantee the above approximation, one can define a set of scores, termed as sensitivities \cite{langberg2010universal} corresponding to each point. This can be used to create coresets via importance sampling. The sensitivity of a point $ \*x $ is defined as $s_{\*x} = \sup_{\*q \in \*Q} \frac{ \mathnormal{f}_{\*q}(\*x)}{\sum_{\*x' \in \*X} \mathnormal{f}_\*q(\*x')}$. Langberg et.al \cite{langberg2010universal} show that using any upper bounds to the sensitivity scores, we can create a probability distribution, which can be used to sample a coreset. The size of the coreset depends on the sum of these upper bounds and the dimension of the query space.

We use the following definitions and inequalities to prove our guarantees.
\begin{theorem}{\label{thm:bernstein}}
\textbf{(Bernstein \cite{dubhashi2009concentration})} Let the scalar random variables $x_{1}, x_{2}, _{\cdots}, x_{n}$ be independent that satisfy $\forall i \in [n]$,  
$\vert x_{i}-\~E[x_{i}]\vert \leq b$. 
Let $X = \sum_{i} x_{i}$ and let $\sigma^{2} = \sum_{i} \sigma_{i}^{2}$ be the variance of $X$, where $\sigma_{i}^{2}$ is the variance of $x_{i}$. 
Then for any $t>0$,
\begin{center}
 $\mbox{Pr}\big(X > \~E[X] + t\big) \leq \exp\bigg(\frac{-t^{2}}{2\sigma^{2}+bt/3} \bigg)$
\end{center}
\end{theorem}
\begin{theorem}{\label{thm:matrixBernstein}}
 \textbf{(Matrix Bernstein \cite{tropp2015introduction})} Let $\*X_{1},\ldots,\*X_{n}$ are independent $d \times d$ random matrices such that $\forall i \in [n], |\|\*X_{i}\| - \~E[\|\*X_{i}\|]| \leq b$ and $\mbox{var}(\|\*X\|) \leq \sigma^{2}$ where $\*X = \sum_{i=1}^{n}\*X_{i}$, then for some $t>0$,
 $$\mbox{Pr}(\|\*X\| - \~E[\|\*X\|] \geq t) \leq d\exp\bigg(\frac{-t^{2}/2}{\sigma^{2}+bt/3}\bigg)$$
\end{theorem}
\begin{definition}{\label{argument:epsNet}} 
\textbf{($\epsilon$-net \cite{haussler1987e})} Given some metric space $\*Q$ its subset $\*P$, i.e., $\*P \subset \*Q$ is an $\epsilon$-net of $\*Q$ on $\ell_{p}$ norm if, $\forall \*x \in \*Q, \exists \*y \in \*P$ such that $\|\*x - \*y\|_{p} \leq \epsilon$.
\end{definition}
The $\epsilon$-net is used to ensure our guarantee for all query vector $\*x$ from a fixed dimensional query space $\*Q$ using union bound argument. Similar argument is used and discussed for various applications \cite{woodruff2014sketching}. 
\section{Related Work}{\label{sec:related}}
Coresets are small summaries of data which 
can be used as a proxy to the original data with provable guarantees. The term was first introduced in \cite{agarwal2004approximating}, where they used coresets for the shape fitting problem. Coresets for clustering problems were described in~\cite{har2004coresets}. 
Feldman et al. \cite{feldman2011unified} gave a generalized framework to construct coresets based on importance sampling using sensitivity scores introduced in \cite{langberg2010universal}. Interested reader can check \cite{woodruff2014sketching, braverman2016new, bachem2017practical}. Various online sampling schemes for spectral approximation are discussed in~\cite{cohen2016online, cohen2017input}. 

Tensor decomposition is unique under minimal assumptions \cite{kruskal1977three}. Therefore it has become very popular in various latent variable modeling applications \cite{anandkumar2014tensor, anandkumar2012method, hsu2012spectral}, learning network parameter of neural networks \cite{janzamin2015beating} etc. However, in general (i.e., without any assumption), most of the tensor problems, including tensor decomposition, are NP-hard \cite{hillar2013most}. There has been much work on fast tensor decomposition techniques. Tensor sketching methods for tensor operations are discussed in \cite{wang2015fast}. They show that by applying FFT to the complete tensor during power iteration, one can save both time and space. This scheme can be used in combination with our scheme. 
A work on element-wise tensor sampling \cite{bhojanapalli2015new} gives a distribution on all the tensor elements and samples a few entries accordingly. For $3$-order, orthogonally decomposable tensors, \cite{song2016sublinear} gives a sub-linear time algorithm for tensor decomposition, which requires the knowledge of norms of slices of the tensor. 
The area of online tensor power iterations has also been explored in \cite{huang2015online, wang2016online}. 
Various heuristics for tensor sketching as well as RandNLA techniques \cite{woodruff2014sketching} over matricized tensors for estimating low-rank tensor approximation have been studied in \cite{song2019relative}. There are few algorithms that use randomized techniques to make CP-ALS, i.e., CP tensor decomposition based on alternating least square method more practical \cite{battaglino2018practical, erichson2020randomized}. Here the author shows various randomized techniques based on sampling and projection to improve the running time and robustness of the CP decomposition. Erichson et al. \cite{erichson2020randomized}, also show that their randomized projection based algorithm can also be used in power iteration based tensor decomposition. For many of these decomposition techniques, our algorithm can be used as a prepossessing. 

In the online setting, for a matrix $\*A \in \~R^{n \times d}$ where rows are coming in streaming manner, the guarantee achieved by \cite{cohen2016online} while preserving additive error spectral approximation $|\|\*A\*x\|^{2} - \|\*C\*x\|^{2}| \leq \epsilon \|\*A\*x\|^{2} + \delta, \forall \*x \in \~R^{d}$, with sample size $O(d(\log d)(\log \epsilon\|\*A\|^{2}/\delta))$.

The problem of $\ell_{p}$ subspace embedding has been explored in both offline~\cite{dasgupta2009sampling, cohen2015p, clarkson2016fast, woodruff2013subspace} and streaming setting~\cite{dickens2018leveraging}. As any offline algorithm to construct coresets can be used as streaming algorithm~\cite{har2004coresets}, we use the known offline algorithms and summarize their results in the streaming version in table \ref{tab:compare}.
Dasgupta et al. \cite{dasgupta2009sampling} show that one can spend $O(nd^{5}\log n)$ time to sample $O(\frac{d^{p+1}}{\epsilon^{2}})$ rows to get a guaranteed $(1\pm \epsilon)$ approximate subspace embedding for any $p$. 
The algorithm in~\cite{woodruff2013subspace} samples $O(n^{1-2/p}\mbox{poly}(d))$ rows and gives $\mbox{poly}(d)$ error relative subspace embedding but in $O(\mbox{nnz}(\*A))$ time. For streaming $\ell_{p}$ subspace embedding~\cite{dickens2018leveraging}, give a one pass deterministic algorithm for $\ell_{p}$ subspace embedding for $1\leq p\leq \infty$. 
For some constant $\gamma \in (0,1)$ the algorithm takes $O(n^{\gamma}d)$ space and $O(n^{\gamma}d^{2}+n^{\gamma}d^{5}\log n)$ update time to return a $1/d^{O(1/\gamma)}$ error relative subspace embedding for any $\ell_{p}$ norm. 
This, however, cannot be made into a constant factor approximation with a sub-linear sample size. We propose various streaming algorithms that give a guaranteed $(1\pm\epsilon)$ relative error approximation for $\ell_{p}$ subspace embedding. 
\section{Algorithms and Guarantees}{\label{sec:algorithms}}
In this section we propose all the six streaming algorithms which are based on two major contributions. 
We first introduce the two algorithmic modules--\online~and \kernelfilter. For real value $p \geq 2$, \online, on arrival of each row, simply decides whether to sample it or not. The probability of sampling is computed based on the stream seen till now,
where as \kernelfilter~works for integer value $p \geq 2$, for every incoming row $\*a_i$, the decision of sampling it, depends on two rows $\grave{\*a}_{i}$ and $\acute{\*a}_{i}$ we define from $\*a_{i}$ such that: for any vector $\*x$, there is a similar transformation $(\grave{\*x}$ and $\acute{\*x})$ and we get, $|\*a_{i}^{T}\*x|^{p}=|\grave{\*a}_{i}^{T}\grave{\*x}|\cdot|\acute{\*a}_{i}^{T}\acute{\*x}|$. For even value $p$ we define $|\*a_{i}^{T}\*x|^{p}=|\acute{\*a}_{i}^{T}\acute{\*x}|^{2}$ and for odd value $p$ we define $|\*a_{i}^{T}\*x|^{p}=|\acute{\*a}_{i}^{T}\acute{\*x}|^{2p/(p+1)}$. We call it kernelization. A similar kernelization is also discussed in \cite{schechtman2011tight} for even value $\ell_{p}$ subspace embedding.

Note that both \online~and \kernelfilter~are restricted streaming algorithms in the sense that each row is selected / processed only when it arrives. For the online nature of the two algorithms we use these as modules in order to create the following algorithms 
\begin{enumerate}
    \item \online+\mrlw: Here, the output streams of \online~is fed to a \mrlw, which is a merge-and-reduce based streaming algorithm based on Lewis Weights. Here the \mrlw~outputs the final coreset. 
    \item \online+\kernelfilter: Here, the output streams from \online~is first kernelized. It is then passed to \kernelfilter, which outputs the final coreset.
    \item \online+\mrlw+\kernelfilter: Here, the output of \online~is fed to \mrlw~further its output is first kernelized and passed to \kernelfilter~, which decide whether to sample it in the final coreset or not.
\end{enumerate}
Note that \online+\mrlw~is a streaming algorithm which works for any $p \geq 2$ where as the algorithm \online+\kernelfilter~even works in a restricted streaming setting for integer valued $p \geq 2$. The algorithm \online+\mrlw+\kernelfilter~is a streaming algorithm works for integer value $p \geq 2$. We also propose \mrlf~which is the streaming version of \online.

The algorithms \online~and \kernelfilter~call a function \oscore($\cdot$), which computes a score for every incoming row, and based on the score, the sampling probability of the row is decided. The score depends on the incoming row (say $\*x_{i}$) and some prior knowledge (say $\*M$) of the data, which it has already seen. Here, we define $\*M = \*X_{i-1}^{T}\*X_{i-1}$ and $\*Q$ is its orthonormal column basis. Here $\*X_{i-1}$ represents the matrix with rows $\{\*x_{1},_{\ldots},\*x_{i-1}\}$ which have arrived so far. Now we present \oscore($\cdot$).
\begin{algorithm}[htpb]
\caption{\oscore($\*x_{i}, \*M, \*M_{inv}, \*Q$)}{\label{alg:onineScore}}
\begin{algorithmic}
\IF{$\*x_{i} \in \mbox{column-space}(\*Q)$}
\STATE $\*M_{inv} = \*M_{inv} - \frac{(\*M_{inv})\*x_{i}\*x_{i}^{T}(\*M_{inv})}{1+\*x_{i}^{T}(\*M_{inv})\*x_{i}}$
\STATE $\*M = \*M + (\*x_{i}\*x_{i}^T)$
\ELSE
\STATE $\*M = \*M + \*x_{i}\*x_{i}^{T}; \*M_{inv} = \*M^{\dagger}$
\STATE $\*Q = \mbox{orthonormal-column-basis}(\*M)$
\ENDIF
\STATE $\tilde{e}_{i} = \*x_{i}^T(\*M_{inv})\*x_{i}$
\STATE Return $\tilde{e}_{i},\*M,\*M_{inv},\*Q$
\end{algorithmic}
\end{algorithm}

Here if the incoming row $\*x_{i} \in \~R^{m}$ lies in the subspace spanned by $\*Q$ (i.e., if $\|\*Q\*x_{i}\|=\|\*x_{i}\|$), then the algorithm takes $O(m^{2})$ time as it need not compute $\*M^{\dagger}$ while computing the term $\*M_{inv}$. If $\*x_{i}$ does not lie in the subspace spanned by $\*Q$ then it takes $O(m^{3})$ as the algorithm needs to compute $\*M^{\dagger}$. Here we have used a modified version of Sherman Morrison formula \cite{sherman1950adjustment} to compute $(\*X_{i}^{T}\*X_{i})^{\dagger} = (\*X_{i-1}^{T}\*X_{i-1}+\*x_{i}\*x_{i}^{T})^{\dagger} = (\*M+\*x_{i}\*x_{i}^{T})^{\dagger}$. Note that in our setup $\*M$ need not be full rank, so we use the formula $(\*X_{i}\*X_{i})^{\dagger} = \*M^{\dagger} - \frac{\*M^{\dagger}\*x_{i}\*x_{i}^{T}\*M^{\dagger}}{1+\*x_{i}^{T}\*M^{\dagger}\*x_{i}}$. 
In the following lemma we prove the correctness of the formula.
\begin{lemma}{\label{lemma:modified-SM}}
 Given a rank-k positive semi-definite matrix $\*M \in \~R^{d \times d}$ and a vector $\*x$ such that it completely lies in the column space of $\*M$. Then we have,
 \begin{equation*}
  (\*M + \*x\*x^{T})^{\dagger} = \*M^{\dagger} - \frac{\*M^{\dagger}\*x\*x^{T}\*M^{\dagger}}{1+\*x^{T}\*M^{\dagger}\*x}
 \end{equation*}
\end{lemma}
\begin{proof}
 The proof is in the similar spirit to lemma \ref{lemma:onlineSummationBound}. Consider $[\*V,\Sigma,\*V] = \mbox{SVD}(\*M)$ and since $\*x$ lies completely in the column space of $\*M$, hence $\exists \*y \in \~R^{k}$ such that $\*V\*y = \*x$. Note that $\*V \in \~R^{d \times k}$.
 \begin{eqnarray*}
  (\*M + \*x\*x^{T})^{\dagger} &=& (\*V\Sigma\*V^{T}+\*V\*y\*y^{T}\*V^{T})^{\dagger} \\
  &=& \*V(\Sigma+\*y\*y^{T})^{-1}\*V^{T} \\
  &=& \*V\bigg(\Sigma^{-1} - \frac{\Sigma^{-1}\*y\*y^{T}\Sigma^{-1}}{1+y^{T}\Sigma^{-1}\*y}\bigg)\*V \\
  &=& \*V\bigg(\Sigma^{-1}-\frac{\Sigma^{-1}\*V^{T}\*V\*y\*y^{T}\*V^{T}\*V\Sigma^{-1}}{1+y^{T}\*V^{T}\*V\Sigma^{-1}\*V^{T}\*V\*y}\bigg)\*V\\
  &=& \*M^{\dagger} - \frac{\*M^{\dagger}\*x\*x^{T}\*M^{\dagger}}{1+\*x^{T}\*M^{\dagger}\*x}
 \end{eqnarray*}
 In the above analysis, the first couple of inequalities are by substitution. In the third equality, we use Sherman Morrison formula on the smaller $k \times k$ matrix $\Sigma$ and the rank-1 update $\*y\*y^{T}$.
\end{proof}
\subsection{\online}
Here we present our first streaming algorithm which ensures equation~\eqref{eq:contract} for integer valued $p \geq 2$ and equation~\eqref{eq:lp} for any real $p \geq 2$. The algorithm can also be used in restricted steaming (online) settings where for every incoming row, we get only one chance to decide whether to sample it or not. Due to its nature of filtering out rows, we call it \online~algorithm. The algorithm tries to reduce the variance of the difference between the cost from the original and the sampled term. In order to achieve that, we use sensitivity based framework to decide the sampling probability of each row. The sampling probability of a row is proportional to its sensitivity scores. In some sense, the sensitivity score of a row captures the fact that how much the variance of the difference is going to get affected if that row is not present in the set of sampled rows. In other words how much the cost function would be affected if the row is not sampled in the coreset. We discuss it in detail in section \ref{sec:proofs}.
Here we present the \online~algorithm and its corresponding guarantees. 
\begin{algorithm}[htpb]
\caption{\online~}{\label{alg:onlineCoreset}}
\begin{algorithmic}
\REQUIRE Streaming rows $\*a_{i}^T, i = 1, {}_{\cdots} n, p \geq 2, r > 1$
\ENSURE Coreset $\*C$ satisfying eqn \eqref{eq:contract} and \eqref{eq:lp} w.h.p.
\STATE $\*M = \*M_{inv} = \*0^{d \times d}$, $L=0$, $\*C= \emptyset$
\STATE $\*Q =  \mbox{orthonormal-column-basis}\*M$
\WHILE{current row $\*a_{i}^T$ is not the last row}
\STATE $[\tilde{e}_{i}, \*M, \*M_{inv}, \*Q] = \oscore(\*a_{i},\*M, \*M_{inv}, \*Q$)
\STATE $\tilde{l}_{i} = \min\{i^{p/2-1}(\tilde{e}_{i})^{p/2},1\}; L = L+\tilde{l}_{i}; p_{i} = \min\{r\tilde{l}_{i}/L,1\}$
\STATE Sample $\*a_{i}/\sqrt[p]{p_{i}}$ in $\*C$ with probability $p_{i}$
\ENDWHILE
\STATE Return $\*C$
\end{algorithmic}
\end{algorithm}

Every time a row $\*a_{i} \in \~R^{d}$ comes, the \online~ calls the function~\ref{alg:onineScore} (i.e., \oscore($\cdot$)) which returns a score $\tilde{e}_{i}$. Then \online~computes $\tilde{l}_{i}$, which is an upper bound to its sensitivity score. Based on $\tilde{l}_{i}$ the row's sampling probability is decided. We formally define and discuss sensitivity scores of our problem in section \ref{sec:proofs}.

Now for the \oscore($\cdot$) function there can be at most $d$ occasions where an incoming row is not in the row space of the previously seen rows, i.e., $\*Q$. In these cases \oscore($\cdot$) takes $O(d^{3})$ time and for the other, at least $n-d$, cases by Sherman Morrison formula it takes $O(d^{2})$ time to return $\tilde{e}_{i}$. Hence the entire algorithm just takes $O(nd^{2})$ time. Now we summarize the guarantees of \online~in the following theorem.
\begin{theorem}\label{thm:Online}
Given $\*A \in \~R^{n \times d}$ whose rows are coming in streaming manner, \online~selects a set $\*C$ of size $O\Big(\frac{n^{1-2/p}dk}{\epsilon^{2}}\big(1+\log\|\*A\|-d^{-1}\min_{i} \log \|\*a_{i}\|\big)\Big)$ using both working space and update time $O(d^2)$. Suppose $\*Q$ is a fixed $k$-dimensional subspace, then with probability at least $0.99$, for integer value $p \geq 2, \epsilon > 0$, $\forall \*x \in \*Q$, the set $\*C$ satisfies both $p$-order tensor contraction and $\ell_{p}$ subspace embedding as in equations \eqref{eq:contract} and \eqref{eq:lp} respectively.
\end{theorem}
\online~can also be used to get an $\ell_{p}$ subspace embedding for any real $p \geq 2$. It is worth noting that \online~benefits by taking very less working space and computation time, which are independent of $p$ (order of the tensor) and $n$ (input size). However \online~gives a coreset which is sublinear to input size but as $p$ increases the factor $n^{1-2/p}$ tends to $O(n)$. Hence for higher $p$ the coresets might be as big as the entire dataset. We discuss the proof of the theorem along with its supporting lemma in section \ref{sec:proofs}. Due to the simplicity of the algorithm we present its streaming version \mrlf~in section \ref{sec:streaminglf}. It improves the update time for a cost of higher working space.
\subsection{\online+\mrlw}
Here we present a streaming algorithm which returns a coreset for the same problem with its coreset much smaller than that of \online. First we want to point out that our coresets for tensor contraction i.e., equation \eqref{eq:contract} also preserve $\ell_p$ subspace embedding i.e., equation \eqref{eq:lp}. This is mainly due to two reasons. First is that our coreset is a subsample of original data, and second is because of the way we define our sampling probability. 

For simplicity, we show this relation in the offline setting, where we have access to the entire data $\*A$.
For a matrix $\*A \in \~R^{n \times d}$, we intend to preserve the tensor contraction property as in equation \eqref{eq:contract}. We create $\*C$ by sampling original row vectors $\*a_{i}$ with appropriate scaling. Hence the rows in $\*C$ also retains the actual structures of the original rows in $\*A$.
We analyze the variance of the difference between the tensor contraction from the original and the sampled term, through Bernstein inequality \cite{dubhashi2009concentration} and try to reduce it. 
Here we use sensitivity based framework to decide our sampling probability where we know sensitivity scores are well defined for non negative cost function~\cite{langberg2010universal}. Now with the tensor contraction the problem is that for odd $p$ and for some $\*x$, the cost $(\*a_{i}^{T}\*x)^{p}$ could be negative, for some $i \in [n]$. So for every row $i$ we define the sensitivity score as follows,
\begin{equation}{\label{eqn:sensitivity}}
 s_{i} = \sup_{\*x}\frac{|\*a_{i}^{T}\*x|^{p}}{\sum_{j=1}^{n}|\mathbf{a_{j}}^{T}\*x|^{p}}
\end{equation}
Using Langberg et.al. \cite{langberg2010universal} result, by sampling enough number of rows based on above defined sensitivity scores would preserve $\sum_{i=1}^{n}|\*a_{i}^{T}\*x|^{p} = \|\*A\*x\|_{p}^{p}$. The sampled rows create a coreset $\*C$ which is $\ell_{p}$ subspace embedding, i.e., $\forall \*x$, $|\|\*A\*x\|_{p}^{p} - \|\*C\*x\|_{p}^{p}| \leq \epsilon \|\*A\*x\|_{p}^{p}$. We define and discuss the online version of these scores in section~\ref{sec:proofs} which also preserve tensor contraction as in equation \eqref{eq:contract}.
Sampling based methods used in~\cite{dasgupta2009sampling, cohen2015p, clarkson2016fast} to get a coreset for $\ell_{p}$ subspace embedding also preserve tensor contraction. This is because these sampling based methods reduce the variance of the difference between the cost function from the original and the sampled terms.

We know that any offline algorithm can be made into a streaming algorithm using merge and reduce method~\cite{har2004coresets}. For $p \geq 2$ the sampling complexity of \cite{cohen2015p} is best among all other methods we mentioned. Hence here we use Lewis Weights sampling \cite{cohen2015p} as the offline method along with merge and reduce to convert it into a streaming algorithm which we call \mrlw. The following lemma summarizes the guarantee one gets from \mrlw.
\begin{lemma}\label{lemma:Stream-MR}
 Given a set of $n$ streaming rows $\{\*a_{i}\}$, the \mrlw~returns a coreset $\*C$. For integer $p \geq 2$, a fixed $k$-dimensional subspace $\*Q$, with probability $0.99$ and $\epsilon > 0$, $\forall \*x\in \~R^{d}$, $\*C$ satisfies $p$-order tensor contraction and $\ell_{p}$ subspace embedding as in equations \eqref{eq:contract} and \eqref{eq:lp}.

 It requires $O(d^{p/2})$ amortized update time and uses $O(d^{p/2}\epsilon^{-5}\log^{11} n)$ working space to return a coreset $\*C$ of size $O(d^{p/2}\epsilon^{-5}\log^{10} n)$.
\end{lemma}
\begin{proof}\label{proof:Stream-MR}
 Here the data is coming in streaming sense and it is feed to the streaming version of the algorithm in \cite{cohen2015p}, i.e. \mrlw~for $\ell_{p}$ subspace embedding. We use merge and reduce from \cite{har2004coresets} for streaming data. From the results of \cite{cohen2015p} we know that for a set $\*P$ of size $n$ takes $O(nd^{p/2})$ time to return a coreset $\*Q$ of size $O(d^{p/2}(\log d)\epsilon^{-5})$. Note that for the \mrlw~in section 7 of \cite{har2004coresets} we set $M=O(d^{p/2}(\log d)\epsilon^{-5})$. The method returns $\*Q_{i}$ as the $(1 + \delta_{i})$ coreset for the partition $\*P_{i}$ where $|\*P_{i}|$ is either $2^{i}M$ or $0$, here $\rho_{j} = \epsilon/(c(j+1)^{2})$ such that $1+\delta_{i} = \prod_{j=0}^{i} (1 + \rho_{j}) \leq 1 + \epsilon/2, \forall j \in \lceil \log n \rceil$. Thus we have $|\*Q_{i}|$ is $O(d^{p/2}(\log d)(i+1)^{10}\epsilon^{-5})$. In \mrlw~the method reduce sees at max $\log n$ many coresets $\*Q_{i}$ at any point of time. Hence the total working space is $O(d^{p/2}(\log^{11} n)(\log d)\epsilon^{-5})$. Note that while creating the coreset $\*Q_{i}$ for $\*P_{i}$, the \mrlw~never actually uses the entire $\*P_{i}$ and run offline Lewis Weight based sampling. Instead it uses all $\*Q_{j}$, where $j < i$. Hence the offline method of Lewis Weight based sampling is run over $\cup_{j < i} \*Q_{j}$ which is $O(\*Q_{i})$. Now the amortized time spent per update is,
 \begin{eqnarray*}
  && \sum_{i=1}^{\lceil \log (n/M) \rceil} \frac{1}{2^{i}M}O(|\*Q_{i}|d^{p/2}) \\
  &=& \sum_{i=1}^{\lceil \log (n/M) \rceil} \frac{1}{2^{i}M})(M(i+1)^{4}d^{p/2}) \leq O(d^{p/2})
 \end{eqnarray*}
So the finally the algorithm return $\*Q$ as the final coreset of $O(d^{p/2}(\log^{10} n)(\log d)\epsilon^{-5})$ rows and uses $O(d^{p/2})$ amortized update time.
\end{proof}
It also guarantees $\ell_{p}$ subspace embedding for real $p \geq 2$. Further, note that in this case both update time and working space has dominating term as functions of $d$ and $p$. The coreset size also has major contributing factor which is $\epsilon^{-5}$.

Now we propose our second algorithm, where we feed the output of \online~to \mrlw~method. Here every incoming row is fed to \online, which quickly computes a sampling probability and based on which the row gets sampled. Now, if it gets sampled, then we pass it to the \mrlw~method, which returns the final coreset. The entire algorithm gets an improved {\em amortized} update time compared to \mrlw~and improved sampling complexity compared to \online. We call this algorithm \online+\mrlw~and summarize its guarantees in the following theorem.
\begin{theorem}{\label{thm:improvedStream-MR}}
 Given $\*A \in \~R^{n \times d}$ whose rows are given to \online+\mrlw~in streaming manner. It requires $O(d^{2})$ amortized update time and uses working space of $((1-2/p)^{11}d^{p/2}\epsilon^{-5}\log^{11} n)$ to return a coreset $\*C$ of size $((1-2/p)^{10}d^{p/2}\epsilon^{-5}\log^{10} n)$ such that with at least $0.99$ probability, $\*C$ satisfies both $p$-order tensor contraction and $\ell_{p}$ subspace embedding as in equations \eqref{eq:contract} and \eqref{eq:lp}.
\end{theorem}
\begin{proof}\label{proof:improvedStream-MR}
 Here the data is coming in streaming sense. The first method \online~filters out the rows with small sensitivity scores and only the sampled rows (high sensitivity score) are passed to \mrlw. Here the \online~ensures that \mrlw~only gets $\tilde{O}(n^{1-2/p}d)$, hence the amortized update time is same as that of \online, i.e. $O(d^{2})$. Now similar to the above proof \ref{proof:improvedStream-MR}, by the \mrlw~from section 7 of \cite{har2004coresets} we set $M=O(d^{p/2}(\log d)\epsilon^{-5})$. The method returns $\*Q_{i}$ as the $(1 + \delta_{i})$ coreset for the partition $\*P_{i}$ where $|\*P_{i}|$ is either $2^{i}M$ or $0$, here $\rho_{j} = \epsilon/(c(j+1)^{2})$ such that $1+\delta_{i} = \prod_{j=0}^{i} (1 + \rho_{j}) \leq 1 + \epsilon/2, \forall j \in \lceil \log n \rceil$. Thus we have $|\*Q_{i}|$ is $O(d^{p/2}(\log d)(i+1)^{10}\epsilon^{-5})$. Hence the total working space is $O((1-2/p)^{11}d^{p/2}(\log^{11} n)(\log d)\epsilon^{-5})$. So finally \online+\mrlw~returns a coreset $\*Q$ of $O((1-2/p)^{10}d^{p/2}(\log^{10} n)(\log d)\epsilon^{-5})$ rows.
\end{proof}
This is an improved streaming algorithm which gives the same guarantee as lemma \ref{lemma:Stream-MR} but using very less amortized update time. Hence asymptotically, we get an improvement in the overall run time of the algorithm and yet get a coreset which is smaller than that of \online. Similar to \mrlw, \online+\mrlw~also ensures $\ell_{p}$ subspace embedding for real $p \geq 2$.
It is important to note that we could improve the run time of the streaming result because our \online~can be used in an online manner, which returns a sub-linear size coreset (i.e., $o(n)$) and its update time is less than the amortized update time of \mrlw.
Note that \online+\mrlw~is a streaming algorithm, whereas \online~or the next algorithm that we propose, works even in the restricted streaming setting. 
\subsection{\kernelfilter}
Now we discuss our second module which is also a streaming algorithm for the tensor contraction guarantee as equation \eqref{eq:contract}. First we give a reduction from $p$-order function to $q$-order function, where $q \leq 2$. For even valued $p$, $(\grave{\*x},\grave{\*y})$ are same as $(\acute{\*x},\acute{\*y})$. So we define $|\*x^{T}\*y|^{p} = |\acute{\*x}^{T}\acute{\*y}|^{2}$, same as \cite{schechtman2011tight}. For odd value $p$, $(\grave{\*x},\grave{\*y})$ are not same as $(\acute{\*x},\acute{\*y})$ and we define $|\*x^{T}\*y|^{p} = |\acute{\*x}^{T}\acute{\*y}|^{2p/(p+1)}$. For completeness we state the following lemma for both even and odd value $p$.
\begin{lemma}{\label{lemma:kernel}}
 For an integer value $p \geq 2$, a vector $\*x \in \~R^{d}$ can be transformed to $(\grave{\*x}$ and $\acute{\*x})$ such that for any two d-dimensional vectors $\*x$ and $\*y$ with their similar transformations we get,
 \[|\*x^{T}\*y|^{p} = |\grave{\*x}^{T}\grave{\*y}|\cdot|\acute{\*x}^{T}\acute{\*y}| = 
 \begin{cases}
  |\acute{\*x}^{T}\acute{\*y}|^{2}  & \quad \text{if } p \mbox{ even}\\
  |\acute{\*x}^{T}\acute{\*y}|^{2p/(p+1)}  & \quad \text{if } p\ \mbox{odd}\\
  \end{cases}\]
\end{lemma}
\begin{proof}{\label{proof:kernel}}
The term $|\*x^{T}\*y|^{p}=|\*x^{T}\*y|^{\lfloor p/2 \rfloor}|\*x^{T}\*y|^{\lceil p/2 \rceil}$. We define $|\*x^{T}\*y|^{\lfloor p/2 \rfloor} = |\grave{\*x}_{i}^{T}\grave{\*y}| = |\langle \*x \otimes^{\lfloor p/2 \rfloor}, \*y \otimes^{\lfloor p/2 \rfloor} \rangle|$ and $|\*x^{T}\*y|^{\lceil p/2 \rceil} = |\acute{\*x}_{i}^{T}\acute{\*y}| = |\langle \*x \otimes^{\lceil p/2 \rceil}, \*y \otimes^{\lceil p/2 \rceil} \rangle|$. Here the $\grave{\*x}$ and $\acute{\*x}$ are the higher dimensional representation of $\*x$ and similarly $\grave{\*y}$ and $\acute{\*y}$ are defined from $\*y$. For even valued $p$ we know $\lfloor p/2 \rfloor=\lceil p/2 \rceil$, so for simplicity we write as $|\*x^{T}\*y|^{p/2} = |\acute{\*x}_{i}^{T}\acute{\*y}|$. Hence we get $|\*x^{T}\*y|^{p} = |\langle \*x \otimes^{p/2}, \*y \otimes^{p/2} \rangle|^{2} = |\acute{\*x}^{T}\acute{\*y}|^{2}$ which is same as in \cite{schechtman2011tight}. Here the vector $\acute{\*x}$ is the higher dimensional vector, where $\acute{\*x} = \mbox{vec}(\*x \otimes^{p/2}) \in \~R^{p/2}$ and similarly $\acute{\*y}$ is also defined from $\*y$. Now for odd value of $p$ we have $\grave{\*x} = \mbox{vec}(\*x \otimes^{(p-1)/2}) \in \~R^{(p-1)/2}$ and $\acute{\*x} = \mbox{vec}(\*x \otimes^{(p+1)/2}) \in \~R^{(p+1)/2}$. Similarly $\grave{\*y}$ and $\acute{\*y}$ are defined from $\*y$. Further note that $|\grave{\*x}^{T}\grave{\*y}| = |\acute{\*x}^{T}\acute{\*y}|^{(p-1)/(p+1)}$ which gives $|\*x^{T}\*y|^{p} = |\langle \*x \otimes^{(p-1)/2}, \*y \otimes^{(p-1)/2} \rangle|\cdot|\langle \*x \otimes^{(p+1)/2}, \*y \otimes^{(p+1)/2} \rangle| = |\grave{\*x}^{T}\grave{\*y}|\cdot|\acute{\*x}^{T}\acute{\*y}| = |\acute{\*x}^{T}\acute{\*y}|^{2p/(p+1)}$. 
\[|\*x^{T}\*y|^{p} =
  \begin{cases}
  |\acute{\*x}^{T}\acute{\*y}|^{2}  & \quad \text{for even } p \\
  |\acute{\*x}^{T}\acute{\*y}|^{2p/(p+1)}  & \quad \text{for odd } p
  \end{cases}
\]
\end{proof}
Here the novelty is in the kernelization for the odd value $p$. 

Now we give a streaming algorithm which is in the same spirit of \online. For every incoming row $\*a_{i}$ it computes the sampling probability based on its kernelized row $\acute{\*a}_{i}$ and the counterpart of the previously seen rows. As the row $\acute{\*a}_{i}$ only depends on $\*a_{i}$, this algorithm can also be used in an online setting as well. 
So for every incoming row based on the value of $p$ our algorithm converts the $d$ dimensional vector into its corresponding higher dimensional vectors before deciding its sampling complexity. 
Since we give a sampling based coreset, it retains the structure of the input data. So one need not require to kernelize $\*x$ into its corresponding higher dimensional vector. Instead, one can use the same $\*x$ on the sampled coreset to compute the desired operation. We call it \kernelfilter~and give it as algorithm~\ref{alg:slowOnline}.
\begin{algorithm}[htpb]
\caption{\kernelfilter}{\label{alg:slowOnline}}
\begin{algorithmic}
\REQUIRE Streaming rows $\*a_{1}, \*a_{2}, _{\cdots}, \*a_{n}$, $r>1, p\geq2$
\ENSURE Coreset $\*C$ satisfying eqn \eqref{eq:contract} and \eqref{eq:lp} w.h.p.
\STATE $\acute{\*M} = \acute{\*M}_{inv} = 0^{d^{\lceil p/2 \rceil} \times d^{\lceil p/2 \rceil}}; L=0; \*C= \emptyset$
\STATE $\acute{\*Q} = \mbox{orthonormal-column-basis}(\acute{\*M})$
\WHILE {$i \leq n$}
\STATE $\acute{\*a}_{i} = \mbox{vec}(\*a_{i}\otimes^{\lceil p/2 \rceil})$
\STATE $[\acute{e}_{i}, \acute{\*M}, \acute{\*M}_{\mbox{inv}}, \acute{\*Q}] = \oscore(\acute{\*a}_{i},\acute{\*M},\acute{\*M}_{\mbox{inv}},\acute{\*Q})$
\IF{($p$ is {\em even})}
    \STATE $\tilde{l}_{i} = (\acute{e}_{i})$
\ELSE
    \STATE $\tilde{l}_{i} = (\acute{e}_{i})^{p/(p+1)}$
\ENDIF
\STATE $L = L+\tilde{l}_{i}; p_{i} = \min\{r\tilde{l}_{i}/L,1\}$
\STATE Sample $\*a_{i}/\sqrt[p]{p_{i}}$ in $\*C$ with probability $p_{i}$
\ENDWHILE
\end{algorithmic}
\end{algorithm}
We summarize the guarantees of \kernelfilter~in the following two theorems.
\begin{theorem}{\label{thm:slowOnline}}
 Given $\*A \in \~R^{n \times d}$ whose rows are coming in a streaming manner and an \textbf{even} value $p$, the \kernelfilter~selects a set $\*C$ of size $O\Big(\frac{d^{p/2}k}{\epsilon^{2}}\big(1+p(\log \|\*A\| - d^{-p/2}\min_{i}\log \|\*a_{i}\|\big)\Big)$ using working space and update time $O(d^{p})$. Suppose $\*Q$ is a fixed $k$-dimensional subspace, then with probability at least $0.99, \epsilon > 0, \forall \*x \in \*Q$ we have $\*C$ satisfying both $p$-order tensor contraction and $\ell_{p}$ subspace embedding as equations \eqref{eq:contract} and \eqref{eq:lp} respectively.
\end{theorem}
\begin{theorem}{\label{thm:slowOnlineOdd}}
 Given $\*A \in \~R^{n \times d}$ whose rows are coming in a streaming manner and an \textbf{odd} integer $p$, $p\ge3$, the algorithm \kernelfilter~selects a set $\*C$ of size $O\Big(\frac{n^{1/(p+1)}d^{p/2}k}{\epsilon^{2}}\big(1+(p+1)(\log \|\*A\|-d^{-\lceil p/2 \rceil}\min_{i}\log \|\*a_{i}\|)\big)^{p/(p+1)}\Big)$ using working space and update time $O(d^{p+1})$. Suppose $\*Q$ is a fixed $k$-dimensional subspace, then with probability at least $0.99, \epsilon > 0, \forall \*x \in \*Q$ we have $\*C$ satisfying both $p$-order tensor contraction and $\ell_{p}$ subspace embedding as equations \eqref{eq:contract} and \eqref{eq:lp} respectively.
\end{theorem}
Here the novelty is in the odd order case is the way we kernelize the streaming rows. We kernelize it in such way that when we go from $p$ order terms $|\*a_{i}^{T}\*x|^{p}$ to $q$ order term $|\acute{\*a}_{i}^{T}\acute{\*x}|^{q}$, we ensure $q<2$ and yet it is closest to $2$. Due to this, we get the smallest possible factor of $n$ in the final coreset size.
The working space and the computation time of \kernelfilter~is more than that of \online, i.e., it is a function of $d$ and $p$. However, note that compare to \online, the \kernelfilter~returns an asymptotically smaller coreset. This is because the $\tilde{l}_{i}$ gives a tighter upper bound of the online sensitivity score compared to what \online~gives. The final coreset size from \kernelfilter~has no factor of $n$ for even value $p$, and there is a small factor of $n$ for odd value $p$, which decreases as $p$ increases.
We discuss the proof of the above two theorems along with its supporting lemmas in section \ref{sec:proofs}.
\subsection{\online+\kernelfilter}
Here we briefly sketch our fourth algorithm.
We use our \online~algorithm along with \kernelfilter~to give a streaming algorithm that benefits both in space and time. For every incoming row first the \online~quickly decides its sampling probability and samples according to it which is then passed to \kernelfilter~which returns the final coreset. Now we state the guarantee of \online+\kernelfilter~in the following theorem.
\begin{theorem}{\label{thm:improvedOnlineCoreset}}
 Consider a matrix $\*A \in \~R^{n \times d}$ whose rows are coming one at a time and feed to the algorithm \online+\kernelfilter, which takes $O(d^{2})$ amortized update time and uses $O(d^{p+1})$ working space for odd $p$ and $O(d^{p})$ for even to return $\*C$. Suppose $\*Q$ is a $k$-dimensional query space, such that with at least $0.99$ probability, $\epsilon > 0, \forall \*x \in \*Q$, $\*C$ satisfies both $p$-order tensor contraction and $\ell_{p}$ subspace embedding as equations \eqref{eq:contract} and \eqref{eq:lp} respectively. With $\*a_{\min} = \text{arg}\min_{i}\|\*a_{i}\|$ the size of $\*C$ is as follows for integer $p \geq 2$:
 \begin{itemize}
    \item $p$ even: $O\Big(\frac{d^{p/2}k}{\epsilon^{2}}\big(1+p(\log \|\*A\|-d^{-p/2} \log \|\*a_{\min}\|)\big)\Big)$
    \item $p$ odd: $O\Big(\frac{n^{(p-2)/(p^{2}+p)}d^{p/2+1/4}k^{5/4}}{\epsilon^{2}}\big(1+(p+1)(\log \|\*A\|-d^{-\lceil p/2 \rceil}\log \|\*a_{\min}\|)\big)^{p/(p+1)}\Big)$
 \end{itemize}
\end{theorem}
\begin{proof}
 As every incoming row is first passed through the \online~and the sampled rows are further fed to \kernelfilter, hence by theorem \ref{thm:Online}, \online~passes $\tilde{O}(n^{1-2/p}dk)$ rows to \kernelfilter, for some constant distortion (say $1/2$). Now by theorem \ref{thm:slowOnline} the algorithm \kernelfilter~returns the final coreset of size $O\Big(\frac{n^{(p-2)/(p^2+p)}d^{(p/2+1/4)}k^{5/4}}{\epsilon^{2}}\big(1+(p+1)(\log \|\*A\|-d^{\lceil p/2 \rceil}\min_{i}\log \|\*a_{i}\|)\big)^{p/(p+1)}\Big)$ for odd value $p$. This is because the \kernelfilter~has a factor of $n$ in its coreset size for the odd value $p$. For even value $p$ as the coreset size from \kernelfilter~is independent of $n$, hence the final coreset size in this case is same as what \kernelfilter~returns. The amortized update time is $O(d^2)$, same as the update time of \online~as it sees every incoming row. Further the working space is $O(d^{p+1})$ for odd value $p$ and $O(d^{p})$ for even value $p$, which is same as what \kernelfilter~uses.
 
Note that at $p=5$, $n^{(p-2)/(p^{2}+p)} = n^{1/10}$ and every other integer $p \geq 2$ is $o(n^{1/10})$.
\end{proof}
Further unlike \online~in this algorithm the factor of $n$ gradually decreases with increase in $p$. Note that for even value $p$ \online+\kernelfilter~returns a coreset with smallest sampling complexity.
\subsection{\online+\mrlw+\kernelfilter}
Here we propose the fifth algorithm \online+\mrlw+\kernelfilter, to get a coreset which achieves both better sampling complexity as well as amortized update time for odd value $p$. The benefit of this that \online~quickly returns $\tilde{o}(n)$ size coreset. Now since the expected coreset size still has a factor of $n$ which cannot be removed by \kernelfilter, hence to remove this factor we pass every sampled row from \online~to \mrlw. Finally, we feed the sample from \mrlw~to \kernelfilter~for final coreset. We state the guarantee of the algorithm in the following theorem.
\begin{theorem}{\label{thm:lflwkf}}
  Consider $\*A \in \~R^{n \times d}$ whose rows are coming one at a time. For odd value $p$, the algorithm \online+\mrlw+\kernelfilter~takes $O(d^{2})$ amortized update time and uses $O(d^{p+1})$ working space to return $\*C$ such that with at least $0.99$ probability, $\epsilon > 0, \forall \*x \in \*Q$, $\*C$ satisfies both $p$-order tensor contraction and $\ell_{p}$ subspace embedding as equations \eqref{eq:contract} and \eqref{eq:lp} respectively. The size of $C$ is as follows for integer $p \ge 2$ is $O\Big(\frac{d^{p/2+1/2}k^{5/4}}{\epsilon^{2}}\big(1+(p+1)(\log \|\*A\|-d^{-\lceil p/2 \rceil}\min_{i}\log \|\*a_{i}\|)\big)^{p/(p+1)}\Big)$.
\end{theorem}
\begin{proof}{\label{proof:lflwkf}}
 As every incoming row is first passed through the \online~and the sampled rows are further fed to \mrlw, which then passes its sampled rows to \kernelfilter. Here by theorem \ref{thm:Online} \online~passes $\tilde{O}(n^{1-2/p}dk)$ rows to \mrlw, for some constant distortion (say $8/10$). Now by a similar theorem \ref{thm:improvedStream-MR}, \mrlw~will return a new coreset of size $\tilde{O}(d^{p/2}k)$ which ensures another constant distortion (say $8/10$). Now further when the sampled rows from \mrlw~passed through \kernelfilter~it returns the final coreset of size $O\Big(\frac{d^{(p/2+1/2)}k^{5/4}}{\epsilon^{2}}\big(1+(p+1)(\log \|\*A\|-d^{\lceil p/2 \rceil}\min_{i}\log \|\*a_{i}\|)\big)^{p/(p+1)}\Big)$ for the odd value $p$. The amortized update time is $O(d^2)$ same as the update time of \online~as it sees every incoming rows. Further the working space is $O(d^{p+1})$ which is same as what \kernelfilter~uses.
\end{proof}
\subsection{\mrlf}{\label{sec:streaminglf}}
Here we propose our final algorithm. We give a streaming algorithm which does the same task as \online, but takes $\tilde{O}(d_{\zeta})$ update time, where $d_{\zeta} = \max_{i\leq n}(\mbox{nnz}(\*a_{i})) \leq d$. Given a matrix $\*A \in \~R^{n \times d}$ and $p \geq 2$, note that the offline sensitivity scores for $p$-order tensor contraction or $\ell_{p}$ subspace embedding, $\forall i \in [n]$ is,
\begin{eqnarray*}
s_{i} &=& \sup_{\*x} \frac{|\*a_{i}^{T}\*x|^{p}}{\sum_{j\leq n} |\*a_{j}^{T}\*x|^{p}} \\
&\leq& n^{p/2-1} \*a_{i}^{T}(\*A^{T}\*A)^{\dagger}\*a_{i} \\
&=& n^{p/2-1}\|\*u_{i}\|^{2} \\
&=& l_{i}
\end{eqnarray*}
Here $[\*U,\Sigma,\*V] = \mbox{SVD}(\*A)$ and $\*u_{i}$ is the $i^{th}$ row of $\*U$. The above inequality can be verified using a similar analysis of lemma \ref{lemma:onlineSensitivityBound}. Now with sampling probability $p_{i} = \min\{1,rl_{i}/L\}$ for the $i^{th}$ row, where $L = \sum_{i\leq n} l_{i}$ and $r$ is $O(kL\epsilon^{-2})$, our coreset $\*C$ of size $O(n^{1-2/p}dk\epsilon^{-2})$ achieves both $p$-order tensor contraction and $\ell_{p}$ subspace embedding as in equation \eqref{eq:contract} and \eqref{eq:lp}. The value of $r$ can be verified using a similar analysis of lemma \ref{lemma:onlineGuarantee}.

Note that $\*U$ is the orthonormal column basis of $\*A$. The running time of the algorithm is dominated by the computation time of $\*U$. Clarkson et.al. \cite{clarkson2017low} showed that there is a randomized technique to get an constant approximation of $\|\*u_{i}\|, \forall i \in [n]$. The randomized algorithm takes $O(\mbox{nnz}(\*A)(\log n)+d^{3})$ time. Now we propose a streaming algorithm which uses \cite{har2004coresets} merge-and-reduce method on the above mentioned offline algorithm. We call it \mrlf~and summarize its guarantees in the following theorem.
\begin{theorem}{\label{thm:stream-LF}}
 Given a set of $n$ streaming rows $\{\*a_{i}\}$, the \mrlf~returns a coreset $\*C$. For integer $p \geq 2$, a fixed $k$-dimensional subspace $\*Q$, with probability $0.99$ and $\epsilon > 0$, $\forall \*x\in \~R^{d}$, $\*C$ satisfies $p$-order tensor contraction and $\ell_{p}$ subspace embedding as in equations \eqref{eq:contract} and \eqref{eq:lp}.

 It requires $\tilde{O}(d_{\zeta})$ amortized update time and uses $O(n^{1-2/p}dk\epsilon^{-2}\log^{5} n)$ working space to return a coreset $\*C$ of size $O(n^{1-2/p}dk\epsilon^{-2}\log^{4} n)$.
\end{theorem}
\begin{proof}\label{proof:stream-LF}
 Here the data is coming in streaming manner which is fed to \mrlf. We know that for $\*A \in \~R^{n \times d}$ it takes $O(\mbox{nnz}(\*A)(\log n) + d^{3})$ time to return a coreset $\*Q$ of size $O(n^{1-2/p}dk\epsilon^{-2})$. Note that for the \mrlf~in section 7 of \cite{har2004coresets} we set $M=O(n^{1-2/p}dk\epsilon^{-2})$. The method returns $\*Q_{i}$ as the $(1 + \delta_{i})$ coreset for the partition $\*P_{i}$ where $|\*P_{i}|$ is either $2^{i}M$ or $0$, here $\rho_{j} = \epsilon/(c(j+1)^{2})$ such that $1+\delta_{i} = \prod_{j=0}^{i} (1 + \rho_{j}) \leq 1 + \epsilon/2, \forall j \in \lceil \log n \rceil$. Thus we have $|\*Q_{i}|$ is $O(|\*P_{i}|^{1-2/p}dk(i+1)^{4}\epsilon^{-2})$. In \mrlf~the method reduce sees at max $\log n$ many coresets $\*Q_{i}$ at any point of time. Hence the total working space is $O(n^{1-2/p}dk(\log^{4} n)\epsilon^{-2})$. With an argument similar to \mrlw, here the offline line version of \online~is not run on entire $\*P_{i}$ to get a coreset $\*Q_{i}$. Instead it is run on $\cup_{j \leq i} \*Q_{j}$ which is $O(M(i+1)^{4}d_{\zeta})$, where $d_{\zeta} = \max_{i \leq n} \mbox{nnz}(\*a_{i})$
 Now the amortized time spent per update is,
 \begin{eqnarray*}
  \sum_{i=1}^{\lceil \log (n/M) \rceil} \frac{1}{2^{i}M}O(M(i+1)^{4}d_{\zeta}(\log \*Q_{i}) + d^{3}) \leq O(d_{\zeta}\log n)
 \end{eqnarray*}
So the finally the algorithm uses $O(n^{1-2/p}dk(\log^{5} n)\epsilon^{-2})$ working space and returns $\*Q$ as the final coreset of $O(n^{1-2/p}dk(\log^{4} n)\epsilon^{-2})$ rows and uses $O(d_{\zeta}\log n)$ amortized update time.
\end{proof}
So in all our previous algorithms, wherever the online algorithm (\online) is used in the first phase, one can use the streaming algorithm (\mrlf) and get an improve the amortized update time of $\tilde{O}(d_{\zeta})$ from $O(d^{2})$. Thereby the the algorithms such as \mrlf+\mrlw, \mrlf+\kernelfilter, \mrlf+\mrlw+\kernelfilter~gets an improved amortized update time of $\tilde{O}(d_{\zeta})$ but it uses a working space of $\tilde{O}(n^{1-2/p}dk\epsilon^{-2})$. For the simplicity of \online, we give its streaming version as \mrlf~which has the best amortized update time.
\subsection{$p=2$ case}
In the case of a matrix, i.e., $p=2$ the \online~and \kernelfilter~are just the same. This is because, for every incoming row $\*a_{i}$, the kernelization returns the same row itself. Hence \kernelfilter's sampling process is exactly the same as \online. While we use the sensitivity framework, for $p=2$, our proofs are novel in the following sense:
\begin{enumerate}
 \item When creating the sensitivity scores in the online setting, we do not need
 to use a regularization term as~\cite{cohen2016online}, instead relying on a novel analysis when the matrix is rank deficient. Hence we get a relative error bound without making the number of samples depend on the smallest non zero singular value (which~\cite{cohen2016online} need for online row sampling for matrices).
 \item We do not need to use a martingale based argument, since the sampling probability of a row does not depend on the previous samples.
\end{enumerate}
Our algorithm gives a coreset which preserves relative error approximation (i.e., subspace embedding). Note that lemma 3.5 of \cite{cohen2016online} can be used to achieve the same but it requires the knowledge of $\sigma_{\min}(\*A)$ (smallest singular value of $\*A$). There we need to set $\delta = \epsilon\sigma_{min}(\*A)$ which gives sampling complexity as $O(d(\log d)(\log \kappa(\*A))/\epsilon^{2})$. Our algorithm gives relative error approximation even when $\kappa(\*A) = 1$, which is not clear in \cite{cohen2016online}. 
First we give a corollary stating one would get by following the analysis mentioned above, i.e. by using the scalar Bernstein inequality~\ref{thm:bernstein}.
\begin{corollary}\label{lem:matrixcoreset}
 Given a matrix $\*A \in \~R^{n\times d}$ with rows coming one at a time, for $p=2$ our algorithm uses $O(d^{2})$ update time and samples $O\Big(\frac{d}{\epsilon^{2}}\big(d+d\log\|\*A\|-\min_{i}\log \|\*a_{i}\|\big)\Big)$ rows and preserves the following with probability at least $0.99$, $\forall \*x \in \~R^{d}$
  \begin{equation*}
  (1-\epsilon)\|\*A\*x\|^{2} \leq \|\*C\*x\|^{2} \leq (1+\epsilon)\|\*A\*x\|^{2}
 \end{equation*}
\end{corollary}
Just by using Matrix Bernstein inequality~\cite{tropp2011freedman} we can slightly improve the sampling complexity from factor of $O(d^{2})$ to factor of $O(d\log d)$. For simplicity we modify the sampling probability to $p_{i} = \min\{r\tilde{l}_{i},1\}$ and get the following guarantee.
\begin{theorem}{\label{thm:improvedMatrixCoreset}}
 The above modified algorithm samples $O\Big(\frac{d\log d}{\epsilon^{2}}\big(1+\log\|\*A\|-d^{-1}\min_{i} \log \|\*a_{i}\|\big)\Big)$ rows and preserves the following with probability at least $0.99$, $\forall \*x \in \~R^{d}$
 \begin{align*}
  (1-\epsilon)\|\*A\*x\|^{2} \leq \|\*C\*x\|^{2} \leq (1+\epsilon)\|\*A\*x\|^{2}
 \end{align*}
\end{theorem}
\begin{proof}{\label{proof:improvedMatrixCoreset}}
We prove this theorem in 2 parts. First we show that sampling $\*a_{i}$ with probability $p_{i}=\min\{r\tilde{l}_{i},1\}$ where $\tilde{l}_{i} = \*a_{i}^{T}(\*A_{i}^{T}\*A_{i})^{\dagger}\*a_{i}$ preserves $\|\*C^{T}\*C\| \leq (1\pm \epsilon)\|\*A^{T}\*A\|$. Next we give the bound on expected sample size.

For the $i^{th}$ row $\*a_{i}$ we define, $\*u_{i} = (\*A^{T}\*A)^{-1/2}\*a_{i}$ and we define a random matrix $\*W_{i}$ corresponding to it, 
\[ \*W_{i} =
  \begin{cases}
    (1/p_{i} - 1)\*u_{i}\*u_{i}^{T}  & \quad \text{with probability } p_{i}\\
    -\*u_{i}\*u_{i}^{T} & \quad \text{with probability } (1-p_{i})
  \end{cases}
\]
Now we have,
\begin{eqnarray*}
 \tilde{l}_{i} &=& \*a_{i}^{T}(\*A_{i-1}^{T}\*A_{i-1}+\*a_{i}\*a_{i}^{T})^{\dagger}\*a_{i} \\
 &\geq& \*a_{i}^{T}(\*A^{T}\*A)^{\dagger}\*a_{i} \\
 &=& \*u_{i}^{T}\*u_{i}
\end{eqnarray*}
For $p_{i} \geq \min\{r\*u_{i}^{T}\*u_{i},1\}$, if $p_{i} = 1$, then $\|\*W_{i}\| = 0$, else $p_{i} = r\*u_{i}^{T}\*u_{i} < 1$. So we get $\norm{\*W_{i}} \leq 1/r$. 
Next we bound $\~E[\*W_{i}^{2}]$, which is,
\begin{eqnarray*}
 \~E[\*W_{i}^{2}] &=& p_{i}(1/p_{i}-1)^{2}(\*u_{i}\*u_{i}^{T})^{2}+(1-p_{i})(\*u_{i}\*u_{i}^{T})^{2} \\
 &\preceq& (\*u_{i}\*u_{i}^{T})^{2}/p_{i} \\
 &\preceq& (\*u_{i}\*u_{i}^{T})/r
\end{eqnarray*}
Let $\*W = \sum_{i=1}^{n} \*W_{i}$, then variance of $\|\*W\|$ 
\begin{eqnarray*}
\mbox{var}(\norm{\*W}) &=& \sum_{i=1}^{n}\mbox{var}(\|\*W_{i}\|) \\
&\leq& \sum_{i=1}^{n} \~E[\|\*W_{i}\|^{2}] \\ 
&\leq& \bigg\lVert\sum_{j=1}^{n} \*u_{j}\*u_{j}^{T}/r \bigg\rVert \\
&\leq& 1/r
\end{eqnarray*}
Next by applying matrix Bernstein theorem \ref{thm:matrixBernstein} with appropriate $r$ we get,
\begin{equation*}
 \mbox{Pr}(\norm{\*X}\geq\epsilon) \leq d \exp\bigg(\frac{-\epsilon^{2}/2}{2/r+\epsilon/(3r)}\bigg)\leq 0.01
\end{equation*}
This implies that our algorithm preserves spectral approximation with at least $0.99$ probability by setting $r$ as $O(\log d/\epsilon^{2})$.

Then the expected number of samples to preserve $\ell_{2}$ subspace embedding is $O(\sum_{i=1}^{n}\tilde{l}_{i}(\log d)/\epsilon^{2})$. Now from lemma \ref{lemma:onlineSummationBound} we know that for $p=2, \sum_{i=1}^{n}\tilde{l}_{i}$ is $O(d(1+\log \|\*A\|) - \min_{i} \|\*a_{i}\|)$. Finally to get $\mbox{Pr}(\norm{\*W}\geq\epsilon) \leq 0.01$ the algorithm samples $O\Big(\frac{d\log d}{\epsilon^{2}}\big(1+\log\|\*A\| - d^{-1}\min_{i}\log \|\*a_{i}\|\big)\Big)$ rows.
\end{proof}
\section{Proofs}{\label{sec:proofs}}
In this section we prove our main theorems. While doing so whenever needed we also state and prove the supporting lemmas for them. 
\subsection{\online}
Here we give a sketch of the proof for theorem \ref{thm:Online}. For ease of notation, the rows are considered numbered according to their order of arrival. The supporting lemmas are for the online setting, which also works for the streaming case.
We show that it is possible to generalize the notion of sensitivity for the online setting as well as give an upper bound to it. We define the {\em online sensitivity} of any $i^{th}$ row $\*a_{i}^{T}$ as: 
\begin{equation*}
 \sup_{\*x\in \*Q}\frac{|\*a_{i}^T\*x|^{p}}{\sum_{j=1}^{i}|\*a_{j}^T\*x|^{p}} 
\end{equation*}
which can also be redefined as:
\begin{equation*}
 \sup_{\*y\in \*Q'}\frac{|\*u_{i}^T\*y|^{p}}{\sum_{j=1}^{i}|\*u_{j}^T\*y|^{p}}
\end{equation*}
where $\*Q'= \{\*y| \*y = \Sigma \*V^T \*x, \*x\in \*Q\}$, svd$(\*A) = \*U\Sigma \*V^{T}$ and $\*Q$ is the query space. Here $\*u_{i}^{T}$ is the $i^{th}$ row of $\*U$. Notice that the denominator now contains a sum only over the rows that have arrived. 
We note that while online sampling results often need the use of martingales as an analysis tool, e.g.,~\cite{cohen2016online}, in our setting, the sampling probability of each row does depend on the previous rows, but not on whether they were sampled or not. So, the sampling decision
of each row is independent. Hence, the application of Bernstein's inequality \ref{thm:bernstein}
suffices.

We first show that the $\tilde{l}_{i}$, as defined in \online~used to compute the sampling probability $p_{i}$, are upper bounds to the online sensitivity scores. 
\begin{lemma}{\label{lemma:onlineSensitivityBound}}
 Consider $\*A \in \~R^{n \times d}$, whose rows are provided in a streaming manner to \online. Let $\tilde{l}_i = \min\{i^{p/2-1}(\*a_{i}^T\*M^{\dagger}\*a_{i})^{p/2},1\}$, and $\*M$ is a $d \times d$ matrix maintained by the algorithm. Then $\forall i \in [n]$,  $\tilde{l}_i$ satisfies the following,
 \begin{equation*}
  \tilde{l}_i \ge \sup_{\*x}\frac{|\*a_{i}^T\*x|^{p}}{\sum_{j=1}^{i}|\*a_{j}^T\*x|^{p}}
 \end{equation*}
\end{lemma}
\begin{proof}{\label{proof:onlineSensitivityBound}}
We define the restricted streaming (online) sensitivity scores $\tilde{s}_{i}$ for each row $i$ as follows,
\begin{eqnarray*}
 \tilde{s}_{i} &=& \sup_{\*x}\frac{\vert \*a_{i}^{T}\*x\vert^{p}}{\sum_{j=1}^{i}\vert \*a_{j}^{T}\*x\vert^{p}} \\
 &=& \sup_{\*y}\frac{\vert \*u_{i}^{T}\*y\vert^{p}} {\sum_{j=1}^{i}\vert \*u_{j}^{T}\*y\vert^{p}}
\end{eqnarray*}
Here $\*y = \Sigma \*V^{T}\*x$ where $[\*U,\Sigma,\*V] = \mbox{svd}(\*A)$ and $\*u_{i}^{T}$ is the $i^{th}$ row of $\*U$. Now at this $i^{th}$ step we also define $[\*U_{i},\Sigma_{i},\*V_{i}] = \mbox{svd}(\*A_{i})$. So with $\*y = \Sigma_{i} \*V_{i}^{T}\*x$ and $\tilde{\*u}_{i}^{T}$ is the $i^{th}$ row of $\*U_{i}$ we rewrite the above optimization function as follows,
\begin{eqnarray*}
 \tilde{s}_{i} &=& \sup_{\*x}\frac{\vert \*a_{i}^{T}\*x\vert^{p}}{\sum_{j=1}^{i}\vert \*a_{j}^{T}\*x\vert^{p}} \\
 &=& \sup_{\*y}\frac{\vert \tilde{\*u}_{i}^{T}\*y\vert^{p}}{\norm{\*U_{i}\*y}_{p}^{p}} \\ 
 &=& \sup_{\*y}\frac{\vert \tilde{\*u}_{i}^{T}\*y\vert^{p}}{\vert \tilde{\*u}_{i}^{T}\*y\vert^{p}+\sum_{j=1}^{i-1}\vert \tilde{\*u}_{j}^{T}\*y\vert^{p}}
\end{eqnarray*}
Let there be an $\*x^{*}$ which maximizes $\tilde{s}_{i}$. Corresponding to it we have  $\*y^{*} = \Sigma_{i} \*V_{i}^{T}\*x^{*}$. For a fixed $\*x$, let $f(\*x) = \frac{\vert \*a_{i}^{T}\*x\vert^{p}}{\sum_{j=1}^{i}\vert \*a_{j}^{T}\*x\vert^{p}} = \frac{|\*a_{i}^{T}\*x|^{p}}{\|\*A_{i}\*x\|_{p}^{p}}$ and $g(\*y) = \frac{\vert \tilde{\*u}_{i}^{T}\*y\vert^{p}}{\norm{\*U_{i}\*y}_{p}^{p}}$. By assumption we have $f(\*x^{*}) \geq f(\*x), \forall \*x$. 

We prove this by contradiction that $\forall \*y, g(\*y^{*}) \geq g(\*y)$, where $\*y = \Sigma_{i} \*V_{i}^{T}\*x$. Let $\exists \*y'$ such that $g(\*y') \geq g(\*y^{*})$. Then we get $\*x' = \*V_{i}\Sigma_{i}^{-1}\*y'$ for which $f(\*x')\geq f(\*x^{*})$, as by definition we have $f(\*x) = g(\*y)$ for $y = \Sigma_{i}\*V_{i}^{T}\*x$. This contradicts our assumption, unless $\*x' = \*x^{*}$. 

Now to maximize the score, $\tilde{s}_{i}, \*x$ is chosen from the row space of $\*A_{i}$. Next, without loss of generality we assume that $\norm{\*y} = 1$ as we know that if $\*x$ is in the row space of $\*A_{i}$ then $\*y$ is in the row space of $\*U_{i}$. Hence we get $\norm{\*U_{i}\*y} = \norm{\*y} = 1$.

We break denominator into sum of numerator and the rest, i.e. $\norm{\*U_{i}\*y}_{p}^{p} = \vert \tilde{\*u}_{i}^{T}\*y\vert^{p}+\sum_{j=1}^{i-1}\vert \tilde{\*u}_{j}^{T}\*y\vert^{p}$. 
Consider the denominator term as $\sum_{j=1}^{i-1}|\tilde{\*u}_{j}^{T}\*y|^{p} \geq f(n)\bigg(\sum_{j=1}^{i-1} |\tilde{\*u}_{j}^{T}\*y|^{2}\bigg)$. From this we estimate $f(n)$ as follows,
\begin{eqnarray*}
\sum_{j=1}^{i-1}|\tilde{\*u}_{j}^{T}\*y|^{2} &=& \bigg(\sum_{j=1}^{i-1}|\tilde{\*u}_{j}^{T}\*y|^{2}\cdot 1\bigg) \nonumber \\
&\stackrel{(i)}{\leq}& \bigg(\sum_{j=1}^{i-1}|\tilde{\*u}_{j}^{T}\*y|^{2p/2}\bigg)^{2/p}\bigg(\sum_{j=1}^{i-1}1^{p/(p-2)}\bigg)^{1-2/p} \\
&\stackrel{(ii)}{\leq}& \bigg(\sum_{j=1}^{i-1}|\tilde{\*u}_{j}^{T}\*y|^{p}\bigg)^{2/p}\cdot(i)^{1-2/p}
\end{eqnarray*}
Here equation (i) is by holder's inequality, where we have $2/p + 1 - 2/p = 1$. So we rewrite the above term as $\big(\sum_{j=1}^{i-1}|\tilde{\*u}_{j}^{T}\*y|^{p}\big)^{2/p} \big(i\big)^{1-2/p} \geq \sum_{j=1}^{i-1} |\tilde{\*u}_{j}^{T}\*y|^{2} = 1 - |\tilde{\*u}_{i}^{T}\*y|^{2}$. Now substituting this in equation (ii) we get,
\begin{eqnarray*}
\bigg(\sum_{j=1}^{i-1}|\tilde{\*u}_{j}^{T}\*y|^{p}\bigg)^{2/p} &\geq& \bigg(\frac{1}{i}\bigg)^{1-2/p} (1 - |\tilde{\*u}_{i}^{T}\*y|^{2})\\
\bigg(\sum_{j=1}^{i-1}|\tilde{\*u}_{j}^{T}\*y|^{p}\bigg) &\geq& \bigg(\frac{1}{i}\bigg)^{p/2-1}(1 - |\tilde{\*u}_{i}^{T}\*y|^{2})^{p/2}
\end{eqnarray*}
So we get $\tilde{s}_{i} \leq \sup_{\*y}\frac{\vert \tilde{\*u}_{i}^{T}\*y\vert^{p}}{\vert \tilde{\*u}_{i}^{T}\*y\vert^{p}+(1/i)^{p/2-1}(1-\vert \tilde{\*u}_{i}^{T}\*y\vert^{2})^{p/2}}$. Note that this function increases with value of $|\tilde{\*u}_{i}^{T}\*y|$, which is maximum when $\*y = \frac{\tilde{\*u}_{i}}{\norm{\tilde{\*u}_{i}}}$, which gives,
\begin{equation}
 \tilde{s}_{i} \leq \frac{\norm{\tilde{\*u}_{i}}^{p}}{\norm{\tilde{\*u}_{i}}^{p} + (1/i)^{p/2-1}(1-\norm{\tilde{\*u}_{i}}^{2})^{p/2}} \nonumber
\end{equation}
As we know that a function $\frac{a}{a+b} \leq \min\{1,a/b\}$, so we get $\tilde{l}_{i} = \min\{1,i^{p/2-1}\norm{\tilde{\*u}_{i}}^{p}\}$. Note that $\tilde{l}_{i} = i^{p/2-1}\norm{\tilde{\*u}_{i}}^{p}$ when $\norm{\tilde{\*u}_{i}}^{p} < (1/i)^{p/2-1}$.
\end{proof}
Here the scores are similar to leverage scores \cite{woodruff2014sketching} but due to $p$ order and data point coming in online manner \online~charges an extra factor of $i^{p/2-1}$ for every row. Although we have bound on the $\sum_{i}^{n} \tilde{l}_{i}$ from lemma \ref{lemma:onlineSummationBound}, but this factor can be very huge. As $i$ increases which eventually sets many $\tilde{l}_{i} = 1$. Although the $\tilde{l}_{i}$'s are computed very quickly but the algorithm gives a loose upper bound due to the additional factor of $i^{p/2-1}$. Now with these upper bounds we get the following.
\begin{lemma}{\label{lemma:onlineGuarantee}}
Let $r$ provided to \online~be 
$O(k\epsilon^{-2}\sum_{j=1}^{n}\tilde{l}_{j})$. Let \online~returns a coreset $\*C$. Then with probability at least $0.99$, $\forall \*x \in \*Q$, 
$\*C$ satisfies the $p$-order tensor contraction as in equation \eqref{eq:contract} and $\ell_{p}$ subspace embedding as in equation \eqref{eq:lp}.
\end{lemma}
\begin{proof}{\label{proof:onlineGuarantee}}
 For simplicity, we prove this lemma at the last timestamp $n$. But it can also be proved for any timestamp $t_{i}$, which is why the \online~can also be used in the restricted streaming (online) setting.
 
 Now for a fixed $\*x \in \~R^{d}$ and its corresponding $\*y$, we define a random variables as follows, i.e. the choice \online~has for every incoming row $\*a_{i}^{T}$.
 \[ w_{i} =
  \begin{cases}
    \frac{1}{p_{i}}(\*u_{i}^{T}\*y)^{p}  & \quad \text{with probability } p_{i} \\
    0 & \quad \text{with probability } (1-p_{i})
  \end{cases}
\]
where $\*u_{i}^{T}$ is the $i^{th}$ row of $\*U$ for $[\*U,\Sigma,\*V] = \mbox{svd}(\*A)$ and $\*y = \Sigma\*V^{T}\*x$. Here we get $\~E[w_{i}] = (\*u_{i}^{T}\*y)^{p}$. In our online algorithm we have defined $p_{i} = \min\{r\tilde{l}_{i}/\sum_{j=1}^{i}\tilde{l}_{j},1\}$ where $r$ is some constant. When $p_{i} \leq 1$, we have
\begin{eqnarray*}
 p_{i} &=& r\tilde{l}_{i}/\sum_{j=1}^{i}\tilde{l}_{j} \\
 &\geq& \frac{r|\*u_{i}^{T}\*y|^{p}}{\sum_{j=1}^{i}\tilde{l}_{j}\sum_{j=1}^{i}|\*u_{j}^{T}\*y|^{p}} \\ 
 &\geq& \frac{r|\*u_{i}^{T}\*y|^{p}}{\sum_{j=1}^{n}\tilde{l}_{j}\sum_{j=1}^{n}|\*u_{j}^{T}\*y|^{p}}
\end{eqnarray*}
As we are analysing a lower bound on $p_{i}$ and both the terms in the denominator are positive so we extend the sum of first $i$ terms to all the $n$ terms. Now to apply Bernstein inequality \ref{thm:bernstein} we bound the term $\vert w_{i} - \~E[w_{i}]\vert \leq b$. Consider the two possible cases,
 
 \textbf{Case 1:} When $w_{i}$ is non zero, then $|w_{i} - \~E[w_{i}]| \leq \frac{|\*u_{i}^{T}\*y|^{p}}{p_{i}} \leq \frac{|\*u_{i}^{T}\*y|^{p}(\sum_{j=1}^{n} \tilde{l}_{j})\sum_{j=1}^{n} |\*u_{j}^{T}\*y|^{p}}{r|\*u_{i}^{T}\*y|^{p}} = \frac{(\sum_{j=1}^{n} \tilde{l}_{j})\sum_{j=1}^{n} |\*u_{j}^{T}\*y|^{p}}{r}$. Note for $p_{i}=1, \vert w_{i} - \~E[w_{i}]\vert = 0$.
 
 \textbf{Case 2:} When $w_{i}$ is $0$ then $p_{i} < 1$. So we have $1 > \frac{r\tilde{l}_{i}}{\sum_{j=1}^{i}\tilde{l}_{j}} \geq \frac{r|\*u_{i}^{T}\*y|^{p}}{(\sum_{j=1}^{n} \tilde{l}_{j})\sum_{j=1}^{n}|\*u_{j}^{T}\*y|^{p}}$. So the term $|w_{i}-\~E[w_{i}] | = |\~E[w_{i}]| = |(\*u_{i}^{T}\*y)^{p}| < \frac{(\sum_{j=1}^{n} \tilde{l}_{j})\sum_{j=1}^{n} |\*u_{j}^{T}\*y|^{p}}{r}$. 
 
 So by setting $b=\frac{(\sum_{j=1}^{n} \tilde{l}_{j})\sum_{j=1}^{n} |\*u_{j}^{T}\*y|^{p}}{r}$ we can bound the term $|w_{i} - \~E[w_{i}]|$. Next we bound the variance of the sum, i.e. $\sum_{i=1}^{n} \tilde{l}_{i}$. Let $\sigma^{2} = \mbox{var}\big(\sum_{i=1}^{n} w_{i}\big) = \sum_{i=1}^{n} \sigma_{i}^{2}$, since every incoming rows are independent of each other and here we consider $\sigma_{i}^{2} = \mbox{var}(w_{i})$
 \begin{eqnarray*}
  \sigma^{2} &=& \sum_{i=1}^{n} \~E[w_{i}^{2}] - (\~E[w_{i}])^{2} \\ 
  &\leq& \sum_{i=1}^{n}\frac{|\*u_{i}^{T}\*y|^{2p}}{p_{i}} \nonumber \\
  &\leq& \sum_{i=1}^{n}\frac{|\*u_{i}^{T}\*y|^{2p}(\sum_{k=1}^{n} \tilde{l}_{k})\sum_{j=1}^{n} |\*u_{j}^{T}\*y|^{p}}{r|\*u_{i}^{T}\*y|^{p}} \nonumber \\
  &\leq& \frac{(\sum_{k=1}^{n} \tilde{l}_{k})(\sum_{j=1}^{n} |\*u_{j}^{T}\*y|^{p})^{2}}{r} \nonumber
 \end{eqnarray*}
 Note that $\|\*U\*y\|_{p}^{p} = \sum_{j=1}^{n} |\*u_{j}^{T}\*y|^{p}$. Now in Bernstein inequality we set $t = \epsilon \sum_{j=1}^{n} |\*u_{j}^{T}\*y|^{p}$, we get, 
 \begin{eqnarray*}
  \mbox{Pr}\bigg(|W - \sum_{j=1}^{n} (\*u_{j}^{T}\*y)^{p}| \geq \epsilon \sum_{j=1}^{n} |\*u_{j}^{T}\*y|^{p}\bigg) &\leq& \exp\bigg(\frac{\big(\epsilon \sum_{j=1}^{n} |\*u_{j}^{T}\*y|^{p}\big)^{2}}{2\sigma^{2}+bt/3}\bigg) \\
  &\leq& \exp\Bigg(\frac{-r\epsilon^{2}(\|\*U\*y\|_{p}^{p})^{2}}{(\|\*U\*y\|_{p}^{p})^{2}\sum_{j=1}^{n}\tilde{l}_{j}(2+\epsilon/3)}\Bigg) \\
  &=& \exp\Bigg(\frac{-r\epsilon^{2}}{(2+\epsilon/3)\sum_{j=1}^{n}\tilde{l}_{j}}\Bigg) 
 \end{eqnarray*}
 Now to ensure that the above probability at most $0.01, \forall \*x \in \*Q$ we use $\epsilon$-net argument as in \ref{argument:epsNet} where we take a union bound over $(2/\epsilon)^{k}, \*x$ from the net. Note that for our purpose $1/2$-net also suffices. Hence with the union bound over all $\*x$ in $1/2$-net we need to set $r = \frac{2k\sum_{j=1}^{n}\tilde{l}_{j}}{\epsilon^{2}}$, which is $\Big(\frac{2k\sum_{j=1}^{n}\tilde{l}_{j}}{\epsilon^{2}}\Big)$.
 
 Now to ensure the guarantee for $\ell_{p}$ subspace embedding for any $p \geq 2$ as in equation ~\eqref{eq:lp} one can consider the following form of the random variable, 
 \[ w_{i} =
  \begin{cases}
    \frac{1}{p_{i}}|\*u_{i}^{T}\*y|^{p}  & \quad \text{with probability } p_{i} \\
    0 & \quad \text{with probability } (1-p_{i})
  \end{cases}
 \]
and follow the above proof. Finally by setting $r$ as $O\Big(\frac{k\sum_{j=1}^{n}\tilde{l}_{j}} {\epsilon^{2}}\Big), \forall \*x \in \*Q$ one can get
\begin{equation*}
 \mbox{Pr}\bigg(|\|\*C\*x\|_{p}^{p} - \|\*A\*x\|_{p}^{p}| \geq \epsilon \|\*A\*x\|_{p}^{p}\bigg) \leq 0.01
\end{equation*}
Since for both the guarantees of equation \eqref{eq:contract} and \eqref{eq:lp} the sampling probability of every incoming row is the same, just the random variables are different, hence for integer valued $p \geq 2$ the same sampled rows preserves both tensor contraction as in equation~\eqref{eq:contract} and $\ell_{p}$ subspace embedding as in equation~\eqref{eq:lp}.
\end{proof}
Note that the above analysis can also be used for $\ell_{p}$ subspace embedding with real $p \geq 2$. Now in order to bound the number of samples, we need a bound on the quantity $\sum_{j=1}^{n}\tilde{l}_{j}$. The analysis is novel because of the way we use matrix determinant lemma for a rank deficient matrix, which is further used to get a telescopic sum for all the terms.
The following lemma upper bounds the sum.
\begin{lemma}{\label{lemma:onlineSummationBound}}
 The $\tilde{l}_{i}$ in \online~algorithm which satisfies lemma \ref{lemma:onlineSensitivityBound} and lemma \ref{lemma:onlineGuarantee} has
 $\sum_{i=1}^{n}\tilde{l}_{i}=O(n^{1-2/p}(d+d\log \|\*A\| - \min_{i}\log \|\*a_{i}\|))$.
\end{lemma}
\begin{proof}{\label{proof:onlineSummationBound}}
Recall that $\*A_i$ denotes the $i\times d$ matrix of the first $i$ incoming rows. \online~maintains the covariance matrix $\*M$. At the $(i-1)^{th}$ step we have $\*M = \*A_{i-1}^T\*A_{i-1}$. This is then used to define the score $\tilde{l}_{i}$  for the next step $i$, as $\tilde{l}_i = \min\{i^{p/2-1}\tilde{e}_{i}^{p/2},1\}$, where $\tilde{e}_i = \*a_{i}^T (\*M + \*a_{i} \*a_{i}^{T})^{\dagger} \*a_{i} = \*a_{i}^T (\*A_{i}^T \*A_{i})^{\dagger} \*a_{i}$ and $\*a_{i}^{T}$ is the $i^{th}$ row. 
The scores $\tilde{e}_{i}$ are also called online leverage scores. We first give a bound on $\sum_{i=1}^{n} \tilde{e}_{i}$. A similar bound is given in the online matrix row sampling by~\cite{cohen2016online}, albeit for a regularized version of the scores $\tilde{e}_i$. 
As the rows are coming, the rank of $\*M$ increases 
from $1$ to at most $d$. We say that the algorithm is
in phase-$k$ if the rank of $\*M$ equals $k$. For each phase $k \in [1, d-1]$, let $i_k$ denote the index
where row $\*a_{i_k}$ caused a phase-change in $\*M$ i.e. 
rank of $(\*A_{i_k-1}^{T} \*A_{i_k-1})$ is $k-1$, while rank of $(\*A_{i_k}^{T} \*A_{i_k})$ is $k$. 
For each such $i_k$, the online leverage score $\tilde{e}_{i_k} = 1$, since row $\*a_{i_k}$
does not lie in the row space of $\*A_{i_k-1}$. There
are at most $d$ such indices $i_k$.

We now bound the $\sum_{i\in [i_k, i_{k+1}-1]} \tilde{e}_i$. Suppose the $\mbox{thin-SVD}(\*A_{i_k}^T \*A_{i_k})=\*V\Sigma_{i_k} \*V^T$, all entries in $\Sigma_{i_k}$ being positive. 
Furthermore, for any $i$ in this phase, i.e. for $i\in [i_{k}+1, i_{k+1}-1]$, $\*V$ forms the basis of the row space of $\*A_{i}$. Define
$\*X_{i} = \*V^T (\*A_{i}^T \*A_{i}) \*V$ and the $i^{th}$ row $\*a_{i} = \*V \*b_{i}$. Notice that each $\*X_{i}\in \~R^{k\times k}$, and $\*X_{i_k} = \Sigma_{i_k}$. Also, $\*X_{i_k}$ is positive definite. Now for each $i\in [i_{k}+1, i_{k+1}-1]$, we have $\*X_{i} = \*X_{i-1} + \*b_{i} \*b_{i}^T$.

So we have, $\tilde{e}_{i} = \*a_{i}^T(\*A_{i}^T \*A_{i} )^{\dagger}\*a_{i} = \*b_{i}^T\*V^T(\*V(\*X_{i-1}+\*b_{i} \*b_{i}^T)\*V^T)^{\dagger}\*V \*b_{i}=  \*b_{i}^T(\*X_{i-1}+\*b_{i}\*b_{i}^T)^{\dagger}\*b_{i} = \*b_{i}^T(\*X_{i-1}+\*b_{i}\*b_{i}^T)^{-1}\*b_{i}$
where the last equality uses the invertibility of the matrix. 
%
 Since $\*X_{i-1}$ is not rank deficient so by using matrix determinant lemma~\cite{harville1998matrix} on
 $\mbox{det}(\*X_{i-1}+\*b_{i}\*b_{i}^T)$ we get the following,
 \begin{eqnarray*}
  \mbox{det}(\*X_{i-1}+\*b_{i}\*b_{i}^T) &=& \mbox{det}(\*X_{i-1})(1+\*b_{i}^T(\*X_{i-1})^{-1}\*b_{i}) \\
  &\stackrel{(i)}{\geq}& \mbox{det}(\*X_{i-1})(1+\*b_{i}^T(\*X_{i-1}+\*b_{i} \*b_{i}^T)^{-1}\*b_{i}) \\
  &=& \mbox{det}(\*X_{i-1})(1+\tilde{e}_{i}) \\
  &\stackrel{(ii)}{\geq}& \mbox{det}(\*X_{i-1})\exp(\tilde{e}_{i}/2) \\
  \exp(\tilde{e}_{i}/2) &\leq& \frac{\mbox{det}(\*X_{i-1}+\*b_{i}\*b_{i}^T)}{\mbox{det}(\*X_{i-1})}
 \end{eqnarray*}
 Inequality $(i)$ follows as $\*X_{i-1}^{-1} - (\*X_{i-1} + \*b\*b^T)^{-1}\succeq 0$ (i.e. p.s.d.). Inequality $(ii)$ follows from
 the fact that $1+x \ge \exp(x/2)$ for $x \leq 1$. Now with $\tilde{e}_{i_k} =1$, we analyze the product of the remaining terms of the phase $k$,
 \begin{align*}
  \prod_{i\in[i_k+1, i_{k+1}-1]} \exp(\tilde{e}_{i}/2) &\le \prod_{i\in[i_k+1, i_{k+1}-1]}  \frac{\mbox{det}(\*X_{i})}{\mbox{det}(\*X_{i-1})} \\
  &\le \frac{\mbox{det}(\*X_{i_{k+1}-1})}{\mbox{det}(\*X_{i_k+1})}.
 \end{align*}
 Now by taking the product over all phases we get,
\begin{align*}
 \exp\bigg(\sum_{i\in [1, i_{d}-1]} \tilde{e}_{i}/2\bigg) 
 & = \exp((d-1)/2)\Big(\prod_{k\in [1,d-1]} \prod_{i\in[i_k+1, i_{k+1}-1]} \exp(\tilde{e}_{i}/2)\Big) \\
 & = \exp((d-1)/2)\Big(\prod_{k\in [1,d-1]}\frac{\mbox{det}(\*X_{i_{k+1}-1})}{\mbox{det}(\*X_{i_k+1})}\Big) \\ 
 & = \exp((d-1)/2)\Big(\frac{\mbox{det}(X_{i_{2}-1})}{\mbox{det}(X_{i_1+1})}\prod_{k\in [2,d-1]} \frac{\mbox{det}(\*X_{i_{k+1}-1})}{\mbox{det}(\*X_{i_k+1})}\Big)
 \end{align*}
 Because we know that for any phase $k$ we have $(\*A_{i_{k+1}-1}^T \*A_{i_{k+1}-1}) \succeq (\*A_{i_k+1}^T \*A_{i_k+1})$ so we get,
 $\mbox{det}(\*X_{i_{k+1}-1}) \ge \mbox{det}(\*X_{i_k+1})$. Further between inter phases terms, i.e. between the last term of phase $k-1$ and the second term of phase $k$ we have $\mbox{det}(\*X_{i_k-1}) \leq \mbox{det}(\*X_{i_k+1})$. Note that we independently handle the first term of phase $k$, i.e. phase change term. Hence we get $\exp((d-1)/2)$ as there are $d-1$ many $i$ such that $\tilde{e}_{i} =1$. Due to these conditions the product of terms from $1$ to $i_{d}-1$ yields a telescopic product, which gives,
 \begin{align*}
   \exp\bigg(\sum_{i\in [1, i_{d}-1]} \tilde{e}_{i}/2\bigg) & \le \frac{\exp((d-1)/2)\mbox{det}(\*X_{i_{d}-1})}{\mbox{det}(\*X_{i_1+1})} \\
   & \le \frac{\exp((d-1)/2)\mbox{det}(\*A_{i_d}^T \*A_{i_d})}{\mbox{det}(\*X_{i_1+1})}
 \end{align*}
 Furthermore, we know $\tilde{e}_{i_d}=1$, so for $i\in [i_d, n]$, the matrix $\*M$ is full rank. We follow the same argument as above, and obtain the following,
 \begin{align*}
  \exp\bigg(\sum_{i\in [i_d, n]} \tilde{e}_{i}/2\bigg) & \le \frac{\exp(1/2)\mbox{det}(\*A^T \*A)}{\mbox{det}(\*A_{i_d+1}^T \*A_{i_d+1})} \\
  & \le \frac{\exp(1/2)\|\*A\|^d}{\mbox{det}(\*A_{i_d+1}^T \*A_{i_d+1})} 
 \end{align*}
Let $\*a_{i_1+1}$ be the first non independent incoming row. Now multiplying the above two expressions and taking logarithm of both sides, and accounting for the indices $i_k$ for $k\in[2,d]$ we get,
\begin{align*}
    \sum_{i\le n} \tilde{e}_i & \le d/2 + 2 d\log\|\*A\| - 2\log \|\*a_{i_1+1}\|  \\
    & \le d/2 + 2 d\log\|\*A\| - \min_{i}2\log \|\*a_{i}\|. 
\end{align*}
Now, we give a bound on $\sum_{i=1}^{n} \tilde{l}_{i}$ where $\tilde{l}_{i} = \min\{1,i^{p/2-1}\tilde{e}_{i}^{p/2}\} \leq \min\{1,n^{p/2-1}\tilde{e}_{i}^{p/2}\}$. We consider two cases. When $\tilde{e}_i^{p/2} \geq n^{1-p/2}$ then $\tilde{l}_{i} = 1$, this implies that $\tilde{e}_i \geq n^{2/p-1}$. But we know $\sum_{i=1}^{n} \tilde{e}_i 
\le O(d+d\log\|\*A\|-\min_{i}\log \|\*a_{i})\|$ and hence there are at-most $O(n^{1-2/p}(d+d\log\|\*A\|-\min_{i}\log \|\*a_{i}\|))$ indices with $\tilde{l}_{i} = 1$. Now for the case where $\tilde{e}_{i}^{p/2} < n^{1-p/2}$, we get $\tilde{e}_{i}^{p/2-1} \leq (n)^{(1-p/2)(1-2/p)}$. Then $\sum_{i=1}^{n} n^{p/2-1}\tilde{e}_{i}^{p/2} = \sum_{i=1}^{n} n^{p/2-1}\tilde{e}_{i}^{p/2-1} \tilde{e}_{i} \leq \sum_{i=1}^{n}n^{1-2/p}\tilde{e}_{i}$ is $O(n^{1-2/p}(d+d\log\|\*A\|-\min_{i}\log \|\*a_{i}\|))$.
\end{proof}
With lemmas~\ref{lemma:onlineSensitivityBound}, \ref{lemma:onlineGuarantee} and \ref{lemma:onlineSummationBound} we prove that the guarantee in theorem \ref{thm:Online} is achieved by \online. The bound on space is evident from the fact that we are maintaining the matrix $\*M$ in algorithm which uses $O(d^{2})$ space and returns a coreset of size $O\Big(\frac{n^{1-2/p}dk}{\epsilon^{-2}}\big(1+\log\|\*A\|-d^{-1}\min_{i}\log \|\*a_{i}\|\big)\Big)$. 
\subsection{\kernelfilter}
In this section, we give a sketch of the proof of theorem \ref{thm:slowOnline}. We use sensitivity based framework to decide the sampling probability of each incoming row. The novelty in this algorithm is by reducing the $p$ order operation to a $q$ order, where $q$ is either $2$ or less than but very close to $2$. Now we give bound on sensitivity score of every incoming row.
\begin{lemma}{\label{lemma:slowOnlineSensitivityBound}}
 Consider a matrix $\*A \in \~R^{n \times d}$, where rows are provided to \kernelfilter~in streaming manner. The term $\tilde{l}_{i}$ defined in the algorithm upper bounds the online sensitivity score, i.e. $\forall i \in [n],$ as follows,
 \begin{equation*}
  \tilde{l}_{i} \geq \sup_{\*x}\frac{|\*a_{i}^{T}\*x|^{p}}{\sum_{j=1}^{i}|\*a_{j}^{T}\*x|^{p}}
 \end{equation*}
\end{lemma}
\begin{proof}{\label{proof:slowOnlineSensitivityBound}}
 We define the online sensitivity scores $\tilde{s}_{i}$ for each point $i$ as follows,
  \begin{equation*}
    \tilde{s}_{i} = \sup_{\{\*x\mid\|\*x\|=1\}}\frac{|\*a_{i}^{T}\*x|^{p}}{\|\*A_{i}\*x\|_{p}^{p}}
  \end{equation*}
  Let $\acute{\*A}$ be the matrix where its $j^{th}$ row $\acute{\*a}_{j} = \mbox{vec}(\*a_{j} \otimes^{d^{\lceil p/2 \rceil}}) \in \~R^{d^{\lceil p/2 \rceil}}$. Further let $\acute{\*A}_{i}$ is the corresponding matrix of $\*A_{i} \in \~R^{i \times d}$ which represents first $i$ streaming rows. We define $[\acute{\*U}_{i},\acute{\Sigma}_{i},\acute{\*V}_{i}] = \mbox{svd}(\acute{\*A}_{i})$ such that $\acute{\*a}_{i}^{T} = \acute{\*u}_{i}^{T}\acute{\Sigma}_{i}\acute{\*V}_{i}^{T}$. Now for a fixed $\*x \in \~R^{d}$ its corresponding $\acute{\*x}$ is also fixed in its higher dimension. Here $\acute{\Sigma}_{i}\acute{\*V}_{i}^{T}\acute{\*x}=\acute{\*z}$ from which we define unit vector $\acute{\*y} = \acute{\*z}/\|\acute{\*z}\|$.
  Now for even value $p$, similar to \cite{schechtman2011tight} can easily upper bound the terms $\tilde{s}_{i}$ as follows,
  \begin{eqnarray*}
   \tilde{s}_{i} &=& \sup_{\{\*x\mid\|\*x\|=1\}}\frac{|\*a_{i}^{T}\*x|^{p}}{\|\*A_{i}\*x\|_{p}^{p}} \\
   &\leq& \sup_{\{\acute{\*x}\mid\|\acute{\*x}\|=1\}}\frac{|\acute{\*a}_{i}^{T}\acute{\*x}|^{2}}{\|\acute{\*A}_{i}\acute{\*x}\|^{2}} \\
   &=& \sup_{\{\acute{\*y}\mid\|\acute{\*y}\|=1\}}\frac{|\acute{\*u}_{i}^{T}\acute{\*y}|^{2}}{\|\acute{\*U}_{i}\acute{\*y}\|^{2}} \\
   &\leq& \|\acute{\*u}_{i}\|^{2}
  \end{eqnarray*}
  Here every equality is by substitution from our above mentioned assumptions and the final inequality is well known from \cite{cohen2015uniform}. Hence finally we get $\tilde{s}_{i} \leq \|\acute{\*u}_{i}\|^{2}$ for even value $p$ as defined in \kernelfilter.
  
  Now for odd value $p$ we analyze $\tilde{s}_{i}$ as follows,
 \begin{eqnarray*}
  \tilde{s}_{i} &=& \sup_{\{\*x\mid\|\*x\|=1\}}\frac{|\*a_{i}^{T}\*x|^{p}}{\|\*A_{i}\*x\|_{p}^{p}}\\
  &\stackrel{i}{\leq}& \sup_{\{\acute{\*x}\mid\|\acute{\*x}\|=1\}}\frac{|\acute{\*a}_{i}^{T}\acute{\*x}|^{2p/(p+1)}}{\sum_{j\leq i}|\acute{\*a}_{j}^{T}\acute{\*x}|^{2p/(p+1)}} \\
  &=& \sup_{\{\acute{\*x}\mid\|\acute{\*x}\|=1\}}\frac{|\acute{\*a}_{i}^{T}\acute{\*x}|^{2p/(p+1)}} {\|\acute{\*A}_{i}\acute{\*x}\|_{2p/(p+1)}^{2p/(p+1)}} \\
  &=& \sup_{\{\acute{\*y}\mid\|\acute{\*y}\|=1\}}\frac{|\acute{\*u}_{i}^{T}\acute{\*y}|^{2p/(p+1)}} {\|\acute{\*U}_{i}\acute{\*y}\|_{2p/(p+1)}^{2p/(p+1)}} \\
  &\stackrel{ii}{\leq}& \sup_{\{\acute{\*y}\mid\|\acute{\*y}\|=1\}}\frac{|\acute{\*u}_{i}^{T}\acute{\*y}|^{2p/(p+1)}} {\|\acute{\*U}_{i}\acute{\*y}\|^{2p/(p+1)}} \\ 
  &=& \sup_{\{\acute{\*y}\mid\|\acute{\*y}\|=1\}}|\acute{\*u}_{i}^{T}\acute{\*y}|^{2p/(p+1)} \\
  &=& \|\acute{\*u}_{i}\|^{2p/(p+1)} 
 \end{eqnarray*}
 The inequality (i) is by lemma \ref{lemma:kernel}. Next with similar assumption as above let $[\acute{\*U}_{i},\acute{\*\Sigma}_{i},\acute{\*V}_{i}]=\mbox{svd}(\acute{\*A}_{i})$. The inequality (ii) is because $\|\acute{\*U}_{i}\acute{\*y}\|_{2p/(p+1)} \geq \|\acute{\*U}_{i}\acute{\*y}\|$ and finally we get $\tilde{s}_{i} \leq \|\acute{\*u}_{i}\|^{2p/(p+1)}$ as defined in \kernelfilter~for odd $p$ value. Hence we get $\tilde{s}_{i} \leq \|\acute{\*u}_{i}\|^{q}$, where $q = 2$ for even value $p$ and $q = 2p/(p+1)$ for odd value $p$.
\end{proof}
Unlike \online, the algorithm \kernelfilter~does not use any additional factor of $i$. Hence it gives tighter upper bounds to the sensitivity scores compared to what lemma~\ref{lemma:onlineSensitivityBound} gives. It will be evident when we sum these upper bounds while computing the sampling complexity. Also note that \kernelfilter~applies to integer value $p \geq 2$. Next in the following we show with these $\tilde{l}_{i}$'s what value $r$ we need to set to get the desired guarantee as claimed in theorem \ref{thm:slowOnline} and theorem \ref{thm:slowOnlineOdd}.
\begin{lemma}{\label{lemma:slowOnlineGuarantee}}
 In the \kernelfilter~let $r$ is set as $O(k\sum_{i=1}^{n}\tilde{l}_{i}/\epsilon^{2})$ then for some fixed k-dimensional subspace $\*Q$, the set $\*C$ with probability $0.99$ $\forall \*x \in \*Q$ satisfies $p$-order tensor contraction as in equation \eqref{eq:contract} and $\ell_{p}$ subspace embedding as in equation \eqref{eq:lp}.
\end{lemma}
%
\begin{proof}{\label{proof:slowOnlineGuarantee}}
 For simplicity we prove this lemma at the last timestamp $n$. But it can also be proved for any timestamp $t_{i}$ which is why the \kernelfilter~can also be used in restricted streaming (online) setting. Also for a change we prove our claim for $\ell_{p}$ subspace embedding. Now for some fixed $\*x \in \~R^{d}$ consider the following random variable for every row $\*a_i^{T}$ as,
\[ w_{i} =
  \begin{cases}
    (1/p_{i}-1)|\*a_{i}^{T}\*x|^{p}  & \quad \text{w.p. } p_{i} \\
    -|\*a_{i}^{T}\*x|^{p} & \quad \text{w.p. } (1-p_{i})
  \end{cases}
\]
 Note that $\~E[w_{i}] = 0$. Now to show the concentration of the sum of the expected terms we will apply Bernstein's inequality \ref{thm:bernstein} on $W = \sum_{i=1}^{n} w_{i}$. For this first we bound $|w_{i} - \~E[w_{i}]| = |w_{i}| \leq b$ and then we give a bound on $\mbox{var}(W) \leq \sigma^{2}$. 

 Now for the $i^{th}$ timestamp the algorithm \kernelfilter~defines the sampling probability $p_{i} = \min\{1,r\tilde{l}_{i}/\sum_{j \leq i}\tilde{l}_{j}\}$ where $r$ is some constant. If $p_{i}=1$ then $|w_{i}| = 0$, else if $p_{i} <1$ and \kernelfilter~samples the row then $|w_{i}| \leq |\*a_{i}^{T}\*x|^{p}/p_{i} = |\*a_{i}^{T}\*x|^{p}\sum_{j=1}^{i}\tilde{l}_{j}/(r\tilde{l}_{i}) \leq \|\*A_{i}\*x\|_{p}^{p}|\*a_{i}^{T}\*x|^{p}\sum_{j=1}^{i}\tilde{l}_{j}/(r|\*a_{i}^{T}\*x|^{p}) \leq  \|\*A\*x\|_{p}^{p}\sum_{j=1}^{n}\tilde{l}_{j}/r$. Next when \kernelfilter~does not sample the $i^{th}$ row, it means that $p_{i} < 1$, then we have $1 > r\tilde{l}_{i}/\sum_{j=1}^{i}\tilde{l}_{j} \geq r|\*a_{i}^{T}\*x|^{p}/(\|\*A_{i}\*x\|_{p}^{p}\sum_{j=1}^{i}\tilde{l}_{j}) \geq r|\*a_{i}^{T}\*x|^{p}/(\|\*A\*x\|_{p}^{p}\sum_{j=1}^{n}\tilde{l}_{j})$. Finally we get $|\*a_{i}^{T}\*x|^{p} \leq \|\*A\*x\|_{p}^{p}\sum_{j=1}^{n}\tilde{l}_{j}/r$. So for each $i$ we get $|w_{i}| \leq \|\*A\*x\|_{p}^{p}\sum_{j=1}^{n}\tilde{l}_{j}/r$. 

Next we bound the variance of sum of the random variable, i.e. $W = \sum_{i=1}^{n} w_{i}$. Let, $\sigma^{2} = \mbox{var}(W) = \sum_{i=1}^{n} \mbox{var}(w_{i}) = \sum_{i=1}^{n}\~E[w_{i}^{2}]$ as follows,
\begin{eqnarray*}
 \sigma^{2} &=& \sum_{i=1}^{n} \~E[w_{i}^{2}] \\
 &=& \sum_{i=1}^{n} |\*a_{i}^{T}\*x|^{2p}/p_{i} \\ 
 &\leq& \sum_{i=1}^{n} |\*a_{i}^{T}\*x|^{2p}\sum_{j=1}^{i}\tilde{l}_{j}/(r\tilde{l}_{i}) \\
 &=& \|\*A_{i}\*x\|_{p}^{p}\sum_{i=1}^{n} |\*a_{i}^{T}\*x|^{2p}\sum_{j=1}^{i}\tilde{l}_{j}/(r|\*a_{i}^{T}\*x|^{p}) \\
 &\leq& \|\*A\*x\|_{p}^{2p}\sum_{j=1}^{n}\tilde{l}_{j}/r 
\end{eqnarray*}
 Now we can apply Bernstein \ref{thm:bernstein} to bound the probability $\mbox{Pr}(|W| \geq \epsilon\|\*A\*x\|_{p}^{p})$. Here we have $b = \|\*A\*x\|_{p}^{p}\sum_{j=1}^{n}\tilde{l}_{j}/r, \sigma^{2} = \|\*A\*x\|_{2p}^{p}\sum_{j=1}^{n}\tilde{l}_{j}/r$ and we set $t = \epsilon\|\*A\*x\|_{p}^{p}$, then we get
\begin{eqnarray*}
 \mbox{Pr}(|\|\*C\*x\|_{p}^{p} - \|\*A\*x\|_{p}^{p}| \geq \epsilon\|\*A\*x\|_{p}^{p}) &\leq& \exp\bigg(\frac{-(\epsilon\|\*A\*x\|_{p}^{p})^{2}}{2\|\*A\*x\|_{p}^{2p}\sum_{j=1}^{n}\tilde{l}_{j}/r+\epsilon\|\*A\*x\|_{p}^{2p}\sum_{j=1}^{n}\tilde{l}_{j}/3r}\bigg) \\
 &=& \exp\bigg(\frac{-r\epsilon^{2}\|\*A\*x\|_{p}^{2p}}{(2+\epsilon/3)\|\*A\*x\|_{p}^{2p}\sum_{j=1}^{n}\tilde{l}_{j}}\bigg) \\
 &=& \exp\bigg(\frac{-r\epsilon^{2}}{(2+\epsilon/3)\sum_{j=1}^{n}\tilde{l}_{j}}\bigg)
\end{eqnarray*}
Note that $|W| = |\|\*C\*x\|_{p}^{p} - \|\*A\*x\|_{p}^{p}|$. Now to ensure that the above probability at most $0.01, \forall \*x \in \*Q$ we use $\epsilon$-net argument as in \ref{argument:epsNet} where we take a union bound over $(2/\epsilon)^{k}, \*x$ from the net. Note that for our purpose $1/2$-net also suffices. Hence with the union bound over all $\*x$ in $1/2$-net we need to set $r = O(k\epsilon^{-2}\sum_{j=1}^{n}\tilde{l}_{j})$. 

Now to ensure the guarantee for tensor contraction as equation \eqref{eq:contract} one can define 
 \[ w_{i} =
  \begin{cases}
   (1/p_{i}-1)(\*a_{i}^{T}\*x)^{p}  & \quad \text{w.p. } p_{i}\\
   -(\*a_{i}^{T}\*x)^{p} & \quad \text{w.p. } (1-p_{i})
  \end{cases}
 \]
and follow the above proof. By setting the $r = O(k\epsilon^{-2}\sum_{j=1}^{n}\tilde{l}_{j})$ one can get the following $\forall \*x \in \*Q$,
\begin{equation*}
 \mbox{Pr}\bigg(|\sum_{\tilde{\*a}_{j} \in \*C}(\tilde{\*a}_{j}^{T}\*x)^{p} - \sum_{j=1}^{n} (\*a_{j}^{T}\*x)^{p}| \geq \epsilon \sum_{j=1}^{n} |\*a_{j}^{T}\*x|^{p}\bigg) \leq 0.01
\end{equation*}
One may follow the above proof to claim the final guarantee as in equation \ref{eq:contract} using the same sampling complexity. Again similar to \online~as the sampling probability of the rows are same for both tensor contraction and $\ell_{p}$ subspace embedding, hence the same subsampled rows preserves both the properties as in equation \eqref{eq:contract} and \eqref{eq:lp}.
\end{proof}
Now in order to bound the number of samples, we need a bound on the quantity $\sum_{j=1}^{n}\tilde{l}_{j}$ which we demonstrate in the following lemma.
\begin{lemma}{\label{lemma:slowOnlineSummationBound}}
 Let $\*a_{\min} = \text{arg}\min_{i}\|\*a_{i}\|$ and $\tilde{l}_{i}$'s used in \kernelfilter~which satisfy lemma \ref{lemma:slowOnlineSensitivityBound} and \ref{lemma:slowOnlineGuarantee} has bound for $\sum_{i=1}^{n} \tilde{l}_{i}$ as,
 \begin{itemize}
  \item $p$ even: $O(d^{p/2}(1+p(\log \|\*A\| - d^{-p/2}\log \|\*a_{\min}\|)))$
  \item $p$ odd: $O(n^{1/(p+1)}d^{p/2}(1+(p+1)(\log \|\*A\| - d^{-\lceil p/2\rceil}\log \|\*a_{\min}\|))^{p/(p+1)})$
 \end{itemize}
\end{lemma}
%
\subsubsection{Proof of Lemma \ref{lemma:slowOnlineSummationBound}}
\begin{proof}{\label{proof:slowOnlineSummationBound}}
 Let $\acute{c}_{i} = \|\acute{\*u}_{i}\|$. Now for even value $p$ we have $\sum_{i=1}^{n} \tilde{l}_{i}=\sum_{i=1}^{n}\acute{c}_{i}^{2}$. From lemma \ref{lemma:onlineSummationBound} we get $\sum_{i=1}^{n} \acute{c}_{i}^{2}$ is $O(d^{p/2}(1+\log \|\acute{\*A}\|-d^{-p/2}\min_{i}\log \|\acute{\*a}_{i}\|)$. Now with $[\*u,\*\Sigma,\*V] = \mbox{svd}(\*A)$ for $i^{th}$ row of $\acute{\*A}$ we have $\acute{\*a}_{i}^{T} = \mbox{vec}(\*a_{i}^{T} \otimes^{p/2}) = \mbox{vec}((\*u_{i}^{T}\*\Sigma\*V^{T})^{p/2})$. So we get $\|\acute{\*A}\| \leq \sigma_{1}^{p/2}$. Hence $\sum_{i=1}^{n}\tilde{l}_i$ is $O(d^{p/2}(1+p(\log \|\*A\|-d^{-p/2}\min_{i}\log \|\*a_{i}\|)))$. 
 
 Now for the odd $p$ case $\sum_{i=1}^{n} \tilde{l}_{i}=\sum_{i=1}^{n}\acute{c}_{i}^{2p/(p+1)}$. From lemma \ref{lemma:onlineSummationBound} we get $\sum_{i=1}^{n} \acute{c}_{i}^{2}$ is $O(d^{\lceil p/2\rceil}(1+\log \|\acute{\*A}\|-d^{-\lceil p/2\rceil}\min_{i}\log \|\acute{\*a}_{i}\|)$. Again, with $[\*u,\*\Sigma,\*V] = \mbox{svd}(\*A)$ for $i^{th}$ row of $\acute{\*A}$ we have $\acute{\*a}_{i}^{T} = \mbox{vec}(\*a_{i}^{T} \otimes^{\lceil p/2 \rceil}) = \mbox{vec}((\*u_{i}^{T}\*\Sigma\*V^{T})^{\lceil p/2\rceil})$. So we get $\|\acute{\*A}\| \leq \sigma_{1}^{(p+1)/2}$. Hence $\sum_{i=1}^{n}\acute{c}_i^2$ is $O(d^{\lceil p/2 \rceil}(1+(p+1)(\log \|\*A\|-d^{-\lceil p/2\rceil}\min_{i}\log \|\*a_{i}\|)))$. Now let $\acute{\*c}$ is a vector with each index $\acute{\*c}_{i}$ is defined as above. Then in this case we have $\sum_{i=1}^{n}\tilde{l}_{i} = \|\acute{\*c}\|_{2p/(p+1)}^{2p/(p+1)} \leq n^{1/(p+1)}\|\acute{\*c}\|^{2p/(p+1)}$ which is $O(n^{1/(p+1)}d^{p/2}(1+(p+1)(\log \|\*A\|-d^{-\lceil p/2\rceil}\min_{i}\log \|\*a_{i}\|))^{p/(p+1)})$.
\end{proof}
The proof of the above lemma is similar to that of lemma \ref{lemma:onlineSummationBound}. It implies that the lemma \ref{lemma:slowOnlineSensitivityBound} gives tighter sensitivity bounds compared to lemma \ref{lemma:onlineSensitivityBound} as the factor of $n$ decreases, as $p$ increases. Now with lemmas \ref{lemma:slowOnlineSensitivityBound}, \ref{lemma:slowOnlineGuarantee} and \ref{lemma:slowOnlineSummationBound} we prove that the guarantee in theorem \ref{thm:slowOnline} and theorem \ref{thm:slowOnlineOdd} is achieved by \kernelfilter. The working space bound of $O(d^{p})$ is evident from the fact that the algorithm is maintaining a $d^{p/2} \times d^{p/2}$ matrix for even $p$ and for odd $p$ it maintains $d^{\lceil p/2 \rceil} \times d^{\lceil p/2 \rceil}$ matrix, hence a working space of $O(d^{p+1})$ is needed.
\section{Applications}\label{sec:topic}
Here we show how our methods can also be used for learning latent variable models using tensor factorization. We use a corollary, which summarizes the guarantees we get on latent variables by learning them using tensor factorization on our coreset. We discuss it section \ref{sec:application}.
Note that one can always use an even order tensor for estimating the latent variables in a generative model. It will only increase a factor of $O(d)$ in the working space, but by doing so, it will return a smaller coreset, which will be independent of $n$.
\par{\textbf{Streaming Single Topic Model:}}\label{exp}
Here we empirically show how sampling using \online+\kernelfilter~can preserve tensor contraction as in equation \eqref{eq:contract}. This can be used in single topic modeling where documents are coming in a streaming manner. We compare our method with two other sampling schemes, namely -- \uni~and online leverage scores, which we call \online$(2)$.

Here we use a subset of \textbf{20Newsgroups} data (pre-processed). We took a subset of $10$K documents and considered the $100$ most frequent words. We normalized each document vector, such that its $\ell_{1}$ norm is $1$ and created a matrix $\*A \in \~R^{10\text{K} \times 100}$. We feed its row one at a time to \online+\kernelfilter~with $p=3$, which returns a coreset $\*C$. We run tensor based single topic modeling~\cite{anandkumar2014tensor} on $\*A$ and $\*C$, to return $12$ top topic distributions from both. We take the best matching between empirical topics and estimated topics based on $\ell_{1}$ distance and compute the average $\ell_{1}$ difference between them. Here smaller is better. We run this entire method $5$ times and report the median of their $\ell_{1}$ average differences. Here the coreset sizes are over expectation. We use Tensorlab package \cite{vervliet2016tensorlab} to run our experiments in Matlab R2018b. The codes can be found \href{https://github.com/supratim05/Streaming-Coresets-for-Symmetric-Tensor-Factorization}{here}.

\begin{table}[htbp]
 \caption{Streaming Single Topic Modeling}
 \label{tab:empCompare}
 \vskip 0.1in
 \begin{center}
   \begin{sc}
    \begin{tabular}{|c|c|c|c|}
     \hline
     Sample & \uni & \online$(2)$ & \online\\
     & & & +\kernelfilter \\
     \hline
     50 & 0.5725 & 0.6903 & \textbf{0.5299}  \\
     \hline
     100 & 0.5093 & 0.6385 & \textbf{0.4379} \\
     \hline
     200 & 0.4687 & 0.5548 & \textbf{0.3231} \\
     \hline
     500 & 0.3777 & 0.3992 & \textbf{0.2173} \\
     \hline
     1000 & 0.2548 & 0.2318 & \textbf{0.1292} \\
     \hline
    \end{tabular} 
   \end{sc}
 \end{center}
 \vskip -0.1in
\end{table}
From the table, it can be seen that our algorithm \online+\kernelfilter~performs better compare to both \uni~and \online$(2)$, thus supporting our theoretical claims. 
\subsection{Latent Variable Modeling}{\label{sec:application}}
Under the assumption that the data is generated by some generative model such as Gaussian Mixture model, Topic model, Hidden Markov model etc, one can represent the data in terms of higher order (say $3$) moments as $\widetilde{\mcal T_{3}}$ to realize the latent variables~\cite{anandkumar2014tensor}. Next the tensor is reduced to an orthogonally decomposable tensor by multiplying a matrix called whitening matrix ($\*W \in \~R^{d \times k}$), such that $\*W^{T}\*M_{2}\*W = \*I_{k}$. Here $k$ is the number of number of latent variables we are interested and $\*M_{2} \in \~R^{d \times d}$ is the $2^{nd}$ order moment. Now the reduced tensor $\widetilde{\mcal T}_{r} = \widetilde{\mcal T}_{3}(\*W,\*W,\*W)$ is a $k \times k \times k$ orthogonally decomposable tensor. Next by running robust tensor power iteration (RTPI) on $\widetilde{\mcal T}_{r}$ we get the eigenvalue/eigenvector pair on which upon applying inverse whitening transformation we get the estimated latent factors and its corresponding weights \cite{anandkumar2014tensor}.

Note that we give guarantee over the $d \times d \times d$ tensor where as the main theorem 5.3 of~\cite{anandkumar2014tensor} has conditioned over the smaller orthogonally reducible tensor $\widetilde{\mcal T}_{r} \in \~R^{k \times k \times k}$. Now rephrasing the main theorem 5.1 of \cite{anandkumar2014tensor} we get that the $\|\mcal M_{3} - \widetilde{\mcal T}_{3}\| \leq \varepsilon\|\*W\|^{-3}$ where $\mcal M_{3}$ is the true $3$-order tensor with no noise and $\widetilde{\mcal T}_{3}$ is the empirical tensor that we get from the dataset. 
Now we state the guarantees that one gets by applying the RTPI on our sampled data.
\begin{corollary}{\label{coro:tensorFactors}}
 For a dataset $\*A \in \~R^{n \times d}$ with rows coming in streaming fashion and the algorithm \online+\kernelfilter~returns a coreset $\*C$ which guarantees \eqref{eq:contract} such that if for all unit vector $\*x \in \*Q$, it ensures $\epsilon\sum_{i \leq n} |\*a^{T}\*x|^{3} \leq \varepsilon \|\*W\|^{-3}$. Then applying the RTPI on the sampled coreset $\*C$ returns $k$ eigenpairs $\{\lambda_{i},\*v_{i}\}$ of the reduced (orthogonally decomposable) tensor, such that it ensures $\forall i \in [k]$,
 \begin{equation*}
  \|\*v_{\pi(i)} - \*v_{i}\| \leq 8\varepsilon/\lambda_{i} \qquad \qquad \mbox{and} \quad \qquad |\lambda_{\pi(i)}-\lambda_{i}| \leq 5\varepsilon
 \end{equation*}
\end{corollary}
Here precisely we have $\*Q$ as the column space of the $\*W^{\dagger}$, where $\*W$ is a $d \times k$ dimensional whitening matrix as defined above.
\subsubsection{Tensor Contraction}
Now we show empirically that how coreset from \online+\kernelfilter~preserves $4$-order tensor contraction. We compare our method with two other sampling schemes, namely -- \uni~and \online$(2)$. Here \online$(2)$ is the \online~with $p=2$. 

\textbf{Dataset:} We generated a dataset with $200$K rows in $\~R^{30}$. Each coordinate of the row was set with a uniformly generated scalar in $[0,1]$. Further, each row was normalized to have $\ell_{2}$ norm as $1$. So we get a matrix of size $200\text{K} \times 30$, but we ensured that it had rank $12$. Furthermore, $99.99\%$ of the rows in the matrix spanned only an $8$-dimensional subspace in $\~R^{30}$, and it is orthogonal to a $4$ dimensional subspace from which the remaining $0.01\%$ of the rows were generated. We simulated these rows to come in the online fashion and applied the three sampling strategies. From the coreset returned by these sampling strategies, we generated $4$-order tensors $\mcal{\hat{T}}$, and we also create the tensor $\mcal T$ using the entire dataset. The three sampling strategies are \uni, \online$(2)$ and \online+\kernelfilter are as follows.

\textbf{\uni:} Here, we sample rows uniformly at random. It means that every row has a chance of getting sampled with a probability of $r/n$, i.e., every row has an equal probability of getting sampled. Here the parameter $r$ is used to decide the expected number of samples. Intuitively it is highly unlikely to pick a representative row from a subspace spanned by fewer rows. Hence the coreset from this sampling method might not preserve tensor contraction $\forall \*x \in \*Q$.

\textbf{\online$(2)$:} Here, we sample rows based on online leverage scores $c_{i} = \*a_{i}^{T}(\*A_{i}^{T}\*A_{i})^{-1}\*a_{i}$. We define a sampling probability for an incoming row $i$ as $p_{i} = \min\{1,rc_{i}/(\sum_{j=1}^{i}c_{j})\}$. Here the parameter $r$ is used to decide the expected number of samples. Rows with high leverage scores have higher chance of getting sampled. Though leverage score sampling preserved rank of the the data, but it is not known to preserve higher order moments or tensor contractions.

\textbf{\online+\kernelfilter:} Here, every incoming row is first fed to \online. If it samples the row, then it further passed to \kernelfilter, which decides whether to sample the row in the final coreset or not. In the algorithm \kernelfilter~we set the parameter to get desired expected number of samples.

Now we compare the relative error approximation $|\mcal T(\*x,\*x,\*x,\*x) - \hat{\mcal T}(\*x,\*x,\*x,\*x)|/\mcal T(\*x,\*x,\*x,\*x)$, between three sampling schemes mentioned above. Here $\mcal T(\*x,\*x,\*x,\*x) = \sum_{i=1}^{n}(\*a_{i}^T\*x)^4$ and $\hat{\mcal T}(\*x,\*x,\*x,\*x) = \sum_{\tilde{\*a}_i \in \*C}^{n}(\tilde{\*a}_{i}^T\*x)^4$. In table (\ref{tab:query_set}), $\*Q$ is set of right singular vectors of $\*A$ corresponding to the $5$ smallest singular values. This table reports the relative error approximation $|\sum_{\*x\in[\*Q]}\mcal T(\*x,\*x,\*x,\*x) - \sum_{\*x\in[\*Q]}\hat{\mcal T}(\*x,\*x,\*x,\*x)|/\sum_{\*x\in[\*Q]}\mcal T(\*x,\*x,\*x,\*x)$. The table (\ref{tab:max_variance}) reports for $\*x$ as the right singular vector of the smallest singular value of $\*A$. Here we choose this $\*x$ because this direction captures the worst direction, as in the direction which has the highest variance in the sampled data. For each sampling technique and each sample size, we ran $5$ random experiments and reported the median of the experiments. Here, the sample size are in expectation.
\begin{table}[htbp]
 \caption{Relative error for query $\*x \in \*Q$}
 \label{tab:query_set}
 \vskip 0.1in
 \begin{center}
   \begin{sc}
    \begin{tabular}{|c|c|c|c|}
    \hline
    Sample & \uni & \online$(2)$ & \online\\
     & & & +\kernelfilter \\
    \hline
    200 & 1.1663 & 0.2286 & \textbf{0.1576}  \\
    \hline
    250 & 0.4187 & 0.1169 & \textbf{0.0855} \\
    \hline
    300 & 0.6098 & 0.1195 & \textbf{0.0611} \\
    \hline
    350 & 0.5704 & 0.0470 & \textbf{0.0436} \\
    \hline
    \end{tabular}
   \end{sc}
 \end{center}
 \vskip -0.1in
\end{table}
\begin{table}[htbp]
 \caption{Relative error for query $\*x$ as right singular vector of the smallest singular value}
 \label{tab:max_variance}
 \vskip 0.1in
 \begin{center}
  \begin{sc}
  \begin{tabular}{|c|c|c|c|}
   \hline
   Sample & \uni & \online$(2)$ & \online\\
     & & & +\kernelfilter \\
   \hline
   100 & 1.3584 & 0.8842 & \textbf{0.6879} \\
   \hline
   200 & 0.8886 & 0.5005 & \textbf{0.3952} \\
   \hline
   300 & 0.8742 & 0.4195 & \textbf{0.3696} \\
   \hline
   500 & 0.9187 & 0.3574 & \textbf{0.2000} \\
   \hline
  \end{tabular}
  \end{sc}
 \end{center}
 \vskip -0.1in
\end{table}

\section{Conclusion}
In this work, we presented both online and streaming algorithms to create coresets for tensor and $\ell_{p}$ subspace embedding, and showed their applications in latent factor models. The algorithms either match or improve upon a number of existing algorithms for $\ell_p$ subspace embedding for all integer $p\ge 2$.  The core of our approach is using a combination of a fast online subroutine \online~for filtering out most rows and a more expensive subroutine for better subspace approximation.  
Obvious open questions include extending the techniques to $p=1$ as well as improving the coreset size for \kernelfilter, for odd-$p$. It will also be interesting to explore the connection of \kernelfilter~to Lewis weights \cite{cohen2015p}, since both are different ways of mapping the $\ell_p$ problem to $\ell_2$. It will also be interesting to explore both theoretically and empirically that how the randomized CP decomposition \cite{battaglino2018practical,erichson2020randomized} performs in various latent variable models.

\paragraph{Acknowledgements.} We are grateful to the anonymous reviewers for their helpful feedback.  Anirban  acknowledges the kind support of the N. Rama Rao Chair Professorship at IIT Gandhinagar, the Google India AI/ML award (2020), Google Faculty Award (2015), and CISCO University Research Grant (2016). Supratim acknowledges the kind support of Additional Fellowship from IIT Gandhinagar.

\bibliographystyle{unsrt}
\bibliography{refer}


\end{document}